%% file: main.tex
\UseRawInputEncoding % because problems with accents in refs

\documentclass[10pt]{article}
\usepackage[letterpaper, left=1in, top=1in, right=1in, bottom=1in, verbose, ignoremp]{geometry}

\usepackage{url}\RequirePackage[colorlinks,citecolor=blue, linkcolor=blue,urlcolor = blue]{hyperref}
\usepackage{latexsym,amssymb,amsmath,amsfonts,graphicx,color,fancyvrb,amsthm,enumerate,subcaption,mathrsfs}
\usepackage[longnamesfirst,authoryear,round]{natbib}
\usepackage[dvipsnames]{xcolor}
\usepackage{xy}\xyoption{all} \xyoption{poly} \xyoption{knot}
\usepackage{float}
\usepackage{bm}
\usepackage{bbm}
\usepackage{multicol,multirow}
\usepackage{array}
\usepackage{relsize}
\usepackage{chngcntr}
\usepackage{etoolbox}
\usepackage{caption}
\usepackage{tikz}
\usepackage{standalone}
\usetikzlibrary{decorations.pathreplacing,calc}
\usetikzlibrary{shapes,backgrounds}
\usetikzlibrary{patterns}
\usetikzlibrary{cd}
\usepackage{amsopn}
\usepackage{tabularx}
\usepackage{booktabs}
\usepackage{calc}
\usepackage{algorithm}
\usepackage[noend]{algpseudocode}
\usepackage{hyperref}
\usepackage{mathtools}

\usepackage{tikz-cd} % commutative diagrams
\usepackage{amscd} % commutative diagrams
\usepackage{comment}
\usepackage[capitalize,nameinlink,compress]{cleveref}
\crefname{figure}{Figure}{Figures} 
\crefname{equation}{}{} 
\crefname{assumption}{Assumption}{Assumptions}
\crefname{subsection}{Subsection}{Subsections}
\usepackage{natbib}
\newcounter{cdrow}

\newtheorem*{theorem*}{Theorem}

\newtheorem*{claim*}{Claim}

\theoremstyle{definition}

\newtheorem*{definition*}{Definition}

\theoremstyle{remark}

\newtheorem*{example*}{Example}

\def\log{{\rm log}}

\makeatletter
\newcommand*{\op}{%
  \DOTSB
  \mathop{\vphantom{\bigoplus}\mathpalette\matt@op\relax}%
  \slimits@
}
\newcommand\matt@op[2]{%
  \vcenter{\m@th\hbox{\resizebox{\widthof{$#1\bigoplus$}}{!}{$\boxplus$}}}%
}
\makeatother

\makeatletter
\def\@biblabel#1{}
\makeatother

\makeatletter
\patchcmd{\NAT@citex}
  {\@citea\NAT@hyper@{%
     \NAT@nmfmt{\NAT@nm}%
     \hyper@natlinkbreak{\NAT@aysep\NAT@spacechar}{\@citeb\@extra@b@citeb}%
     \NAT@date}}
  {\@citea\NAT@nmfmt{\NAT@nm}%
   \NAT@aysep\NAT@spacechar\NAT@hyper@{\NAT@date}}{}{}

\patchcmd{\NAT@citex}
  {\@citea\NAT@hyper@{%
     \NAT@nmfmt{\NAT@nm}%
     \hyper@natlinkbreak{\NAT@spacechar\NAT@@open\if*#1*\else#1\NAT@spacechar\fi}%
       {\@citeb\@extra@b@citeb}%
     \NAT@date}}
  {\@citea\NAT@nmfmt{\NAT@nm}%
   \NAT@spacechar\NAT@@open\if*#1*\else#1\NAT@spacechar\fi\NAT@hyper@{\NAT@date}}
  {}{}

\makeatother

\begin{document}
\def\spacingset#1{\renewcommand{\baselinestretch}%
{#1}\small\normalsize} \spacingset{1}
%\spacingset{1.45} % DON'T change the spacing!

\begin{flushleft}
{\Large{\textbf{Tracking Representation Dynamics in Large Language Models with Persistent Homology}}}
\newline
\\
Naman Malhotra$^{1,2}$, Jay Ambadkar$^{1,2}$, Abhinav Gupta$^{1,2}$, Kushal Kasivel$^{1,2}$, Abbas Schwarz$^{1,2}$, Kamillo Ferry$^{1}$, and Anthea Monod$^{1,\dagger}$
\\
\bigskip
\bf{1} Department of Mathematics, Imperial College London, UK
\\
\bf{2} Department of Computing, Imperial College London, UK
\\
 \bigskip
 $\dagger$ Corresponding e-mail: a.monod@imperial.ac.uk
\end{flushleft}

\section*{Abstract}
Large language models are commonly aligned through supervised fine-tuning, yet little is known about how their internal representations evolve during this process. We study alignment dynamics using persistent homology by tracking the topology of activation spaces throughout fine-tuning. Across four transformer language models ranging from 1B to 7B parameters and three alignment objectives corresponding to helpful, harmless, and mixed training data, we find that the majority of topological reorganization occurs during the earliest stages of training. A dense checkpoint analysis reveals a transient peak in topological activity followed by rapid stabilization. We further show that different alignment objectives induce distinguishable topological trajectories, while instruction-tuned and pretrained models exhibit qualitatively different patterns of evolution. Our results suggest that persistent homology provides a complementary perspective on alignment, revealing representation-level changes that are not apparent from behavioral metrics alone.

\paragraph{Keywords:}  Persistent homology, Large language models, Representation geometry, Alignment\\

%%%%%%%%%%%%%%%%%%%%%%%%%%%%%%%%%%%%%%%%%%%%%%%%%%%

\section{Introduction}

Large language models (LLMs) are typically aligned through supervised fine-tuning and preference-based training in order to improve helpfulness, harmlessness, and instruction-following behavior. While the behavioral effects of alignment have been extensively studied, considerably less is known about how the internal representations of a model evolve during the alignment process itself.

Recent work in mechanistic interpretability has shown that meaningful changes in a model's internal state can occur even when behavioral metrics remain unchanged. This phenomenon motivates the search for complementary approaches that characterize representation dynamics directly. Here, we study these dynamics using \emph{persistent homology} (PH), a tool from topological data analysis that extracts multiscale topological structure from high-dimensional point clouds.
Building on recent applications of PH to LLM activations, we investigate how the topology of latent activation spaces evolves during alignment fine-tuning. 

We analyze four transformer models ranging from 1B to 7B parameters 
and track activation clouds across training checkpoints under helpful, harmless, and mixed alignment 
objectives derived from the Anthropic HH-RLHF dataset (Helpful and Harmless Reinforcement Learning from Human Feedback \citep{bai2022training}), 
a standard alignment dataset containing human preference judgments. 
By computing PH at successive stages of training, we obtain a topological view of representation dynamics throughout the alignment process.

\paragraph{Contributions.} Using persistent homology to study activation spaces throughout alignment fine-tuning, we make the following contributions:

\begin{itemize}
\item Across different model sizes, a dense checkpoint analysis reveals a transient peak in topological reorganization during early training, followed by rapid stabilization.

\item Different alignment objectives induce distinguishable topological trajectories despite often producing similar behavioral outcomes.

\item The observed dynamics depend strongly on the model's initial state, with instruction-tuned and pretrained models exhibiting qualitatively different patterns of evolution.

\item We release a fully reproducible implementation of the complete analysis pipeline available at \url{https://github.com/malhotranaman/tracking-representation-dynamics-with-persistent-homology}, from activation extraction and persistent homology computation to statistical evaluation, to facilitate future topological studies of large language models.
\end{itemize}

Together, these findings suggest that persistent homology provides a useful lens for studying representation dynamics during alignment and reveals aspects of fine-tuning that are only partially reflected in behavioral metrics.

%%%%%%%%%%%%%%%%%%%%%%%%%%%%%%%%%%%%%%%%%%%%%%%%%%%

\section{Background}
\label{sec:background}

We briefly review PH and overview the context of understanding representation dynamics and alignment in LLMs.

\subsection{Persistent Homology}

Persistent homology (PH) is a tool from topological data analysis that quantifies the multiscale topology of data. 
Given a point cloud $X$ equipped with a distance metric, PH constructs a nested sequence of simplicial complexes 
called a \emph{filtration} by progressively increasing a scale parameter $\varepsilon \geq 0$. In this work, we use the \emph{Vietoris--Rips} filtration: at scale $\varepsilon$, two points are connected by an edge whenever their distance is at most $\varepsilon$, and higher-dimensional simplices are included whenever all pairwise distances between their vertices are at most $\varepsilon$. As $\varepsilon$ increases, topological features such as connected components and loops appear and disappear. These features are summarized by a \emph{barcode}, which records the birth and death scales $(b_i,d_i)$ of each topological feature.

\begin{figure}[b]
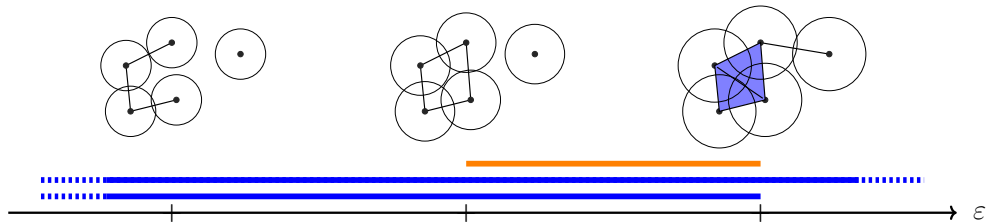

\centering
\includestandalone[width=.80\linewidth]{figures/vietoris-rips-sketch}
\caption{Example of a Vietoris--Rips complex with three values for \(\varepsilon\) and corresponding barcode shown. Blue bars represent connected components while the orange bar refers to a loop.}
\label{fig:vr-example}
\end{figure}

In our setting, PH is applied to activation point clouds derived from latent token representations in LLMs. We restrict attention to homological dimensions $H_0$ and $H_1$, corresponding to connected components and loops, respectively.

\subsection{Related Work} 

LLMs are typically aligned through supervised fine-tuning and preference optimization in order to produce more helpful and harmless behavior \citep{bai2022training,ouyang2022training}. Although alignment often appears to modify only a small subset of model behaviors \citep{zhou2023lima,jain2024mechanistically,lee2024mechanistic,arditi2024refusal}, substantial internal reorganization can occur before these behavioral changes become apparent \citep{barak2022hidden,nanda2023progress}. Understanding the dynamics of internal representations during alignment therefore remains an important open problem.

PH has been used to characterize representations in deep neural networks \citep{naitzat2020topology,rieck2019neural,moor2020topological} and, more recently, in LLMs \citep{gardinazzi2024persistent,fay2026shape,tang2026topological}. In contrast to prior work, which has largely focused on static representations or specific training phenomena, we use PH to study the evolution of representation topology throughout alignment fine-tuning in contemporary LLMs.

%%%%%%%%%%%%%%%%%%%%%%%%%%%%%%%%%%%%%%%%%%%%%%%%%%%

\section{Representation Topology Pipeline}
\label{sec:pipeline}

We track the evolution of latent activation spaces throughout alignment fine-tuning. At multiple training checkpoints, we extract activation point clouds from a fixed evaluation set, compute persistent homology, summarize the resulting barcodes using topological descriptors, and compare their evolution across alignment objectives and model families. 

\paragraph{Models and Alignment Objectives.} We study four decoder-only transformer language models with a range of 1B--7B parameters: 
TinyLlama-1.1B \citep{zhang2024tinyllama}, Gemma-3-1B \citep{team2024gemma}, Phi-3-mini-3.8B (instruction-tuned) \citep{abdin2024phi}, and Mistral-7B (v0.1) \citep{mistral7b}. 

Each model is fine-tuned using Low-Rank Adaptation (LoRA) \citep{hu2022lora}, freezing the pretrained model weights and performing low-rank updates.
This reduces the number of trainable parameters compared to full fine-tuning under 
three alignment objectives derived from the HH-RLHF dataset \citep{bai2022training}. 
In all experiments, the same optimization hyperparameters and random seed are used, only differing in the training objective.

The dataset consists of pairs of candidate assistant responses together with human preference labels. 
We consider three training regimes: \emph{helpful} (H), encouraging informative responses to user queries; 
\emph{harmless} (S), encouraging safe and non-harmful behavior; 
and \emph{mixed} (M), combining both objectives. 
These objectives provide distinct but related alignment signals, allowing us to investigate 
how different forms of alignment influence the topology of latent activation spaces during fine-tuning.

Training checkpoints are saved every 50 optimization steps up to 300 steps for the 1B models, with longer schedules used for larger models. To study the earliest stages of training in greater detail, we additionally perform a dense checkpoint analysis every 5 steps up to step 50 (or the corresponding early-training window for larger models).

\paragraph{Activation Point Clouds and Persistent Homology.} 
At each checkpoint, we evaluate the model on a fixed set of 750 prompts comprising 
250 helpful, 250 harmless, and 250 benign examples drawn from HH-RLHF. 
For each prompt and selected transformer layer, we extract the final-token hidden state 
and treat it as a point in the model's latent representation space. 
Repeating this procedure over all prompts yields an activation point cloud for a given layer and checkpoint. 
Following the methodology from \cite{fay2026shape}, we extract activations from five evenly spaced layers throughout the network depth.

For each activation point cloud, we compute its Vietoris--Rips complex using the Euclidean metric. 
We restrict attention to the 0- and 1-dimensional components ($H_0$ resp.\ $H_1$), corresponding to connected components and loops, respectively. PH was computed using the \textsc{Ripser}~\citep{bauer2021ripser} software.

\paragraph{Topological Summaries.} 
To obtain fixed-length representations suitable for comparison across checkpoints and objectives, 
each barcode is summarized by the same 41 features as in \cite{fay2026shape}.
The summary combines barcode counts, persistence entropy, moments of the birth, death, 
and persistence distributions, and several scale-invariant statistics computed separately for $H_0$ and $H_1$.

%%%%%%%%%%%%%%%%%%%%%%%%%%%%%%%%%%%%%%%%%%%%%%%%%%%

\section{Analysis Framework}
\label{sec:methods}

Each activation point cloud is repeatedly subsampled in order to estimate variability arising from finite prompt sets. Since these subsamples overlap and therefore do not constitute independent observations, all statistical inference is performed at the cell level rather than the subsample level. Throughout, a \emph{cell} denotes a fixed combination of model, objective, evaluation condition, layer, and checkpoint.

\paragraph{Cell Aggregation and Effect Sizes.} 
For each cell, we average the $B=64$ subsample summaries into a single mean vector
$
\bar{\mathbf z}=\frac{1}{B}\sum_{i=1}^{B}\mathbf z_i\in\mathbb R^{41}.
$
A (model, objective, condition, layer) trajectory is then represented by the sequence $\bar{\mathbf z}(t_0),\ldots,\bar{\mathbf z}(t_T)$. Features are standardized within each model prior to multivariate analysis.

We report both univariate and multivariate effect sizes. At the feature level, we use Hedges' $g$ and Cliff's $\delta$ together with percentile-bootstrap 95\% confidence intervals. For the full 41-dimensional representation, we use the \emph{energy distance} \citep{szekely2013energy}, $$\mathcal{E}(A,B)=\big(2\,\overline{\|a-b\|}-\overline{\|a-a'\|}-\overline{\|b-b'\|}\big)^{1/2},$$ which remains stable in settings with relatively few cell-level observations.

\paragraph{Objective Separation.}
% \label{sec:methU}
To quantify separation between alignment objectives, we define the \emph{trajectory dispersion} statistic
\begin{equation}\label{sec:methU}
U:=\sum_t\Big(
|\bar{\mathbf z}_H(t)-\bar{\mathbf z}_M(t)|^2+
|\bar{\mathbf z}_H(t)-\bar{\mathbf z}_S(t)|^2+
|\bar{\mathbf z}_S(t)-\bar{\mathbf z}_M(t)|^2
\Big),
\end{equation}
which measures the aggregate separation of the helpful, harmless, and mixed trajectories.

Significance is assessed via permutation tests over whole trajectories rather than individual subsamples. Because a single trajectory per objective yields a degenerate permutation test, we additionally perform three-seed replications for TinyLlama-1.1B, Gemma-3-1B, and Mistral-7B, producing nine trajectories per model for objective-separation analyses.

\paragraph{Early-Shock Analysis.} \label{sec:methshock}
To quantify topological reorganization, we compute the \emph{Wasserstein} distance between $H_1$ barcodes at consecutive checkpoints, yielding a topological velocity $v_t$. We summarize the \emph{concentration} of topological change by
\begin{equation}\label{eq:concentration-for-shock}
C:=\frac{\sum_{t\le T/3}v_t}{\sum_t v_t},
\end{equation}
the fraction of total topological movement occurring within the first third of training. We use the first third as a coarse, model-agnostic notion of an \emph{early} training phase, allowing comparisons across models with different numbers of checkpoints without committing to a particular change-point or peak location. We deliberately use a broad early-training window rather than the first peak itself, since peak location varies across models and objectives.

To assess whether the observed concentration is genuinely temporal rather than a consequence of total movement, we construct a checkpoint-order null by randomly permuting checkpoint order and recomputing both $v_t$ and $C$. 

\paragraph{Behavioral Comparisons.}
To compare representation-level and behavioral dynamics, we track changes in the model's output distribution throughout fine-tuning. For each checkpoint, we compute the Kullback--Leibler divergence from the initial model and define a behavioral velocity using the Jensen--Shannon divergence between consecutive checkpoints, averaged over a fixed evaluation set. This yields a behavioral trajectory on the same checkpoint axis as the topological velocity, allowing for the comparison of timing of representation and behavioral change.

\paragraph{Controls.}
We perform several control analyses to assess the robustness and specificity of the observed topological phenomena. First, we compare PH against standard geometric summaries computed on the same activation clouds, including centroid drift, total variance, and mean pairwise distance. Second, we compare the observed topological signatures to an isotropic Gaussian baseline matched in dimension and sample size. Finally, we investigate the role of token selection by reconstructing activation clouds from the final 16 token states of each prompt and repeating the corresponding topological analyses. Additional robustness checks are reported in Appendix \ref{app:supp}.

\section{Results and Discussion}
\label{sec:results}

Unless otherwise noted, all analyses use the second-to-last probed layer on the harmful evaluation condition. Features are standardized within each model and permutation tests use at least 2000 randomizations. We organize the results around three research questions: (RQ1) When does topological reorganization occur? (RQ2) How do alignment objectives differ? (RQ3) How do topological and behavioral dynamics relate?

\subsection{Topology Reorganizes Early During Fine-Tuning (RQ1)}
\label{sec:res-shock}

\begin{figure}[tb]
  \centering
  \includegraphics[width=.65\linewidth]{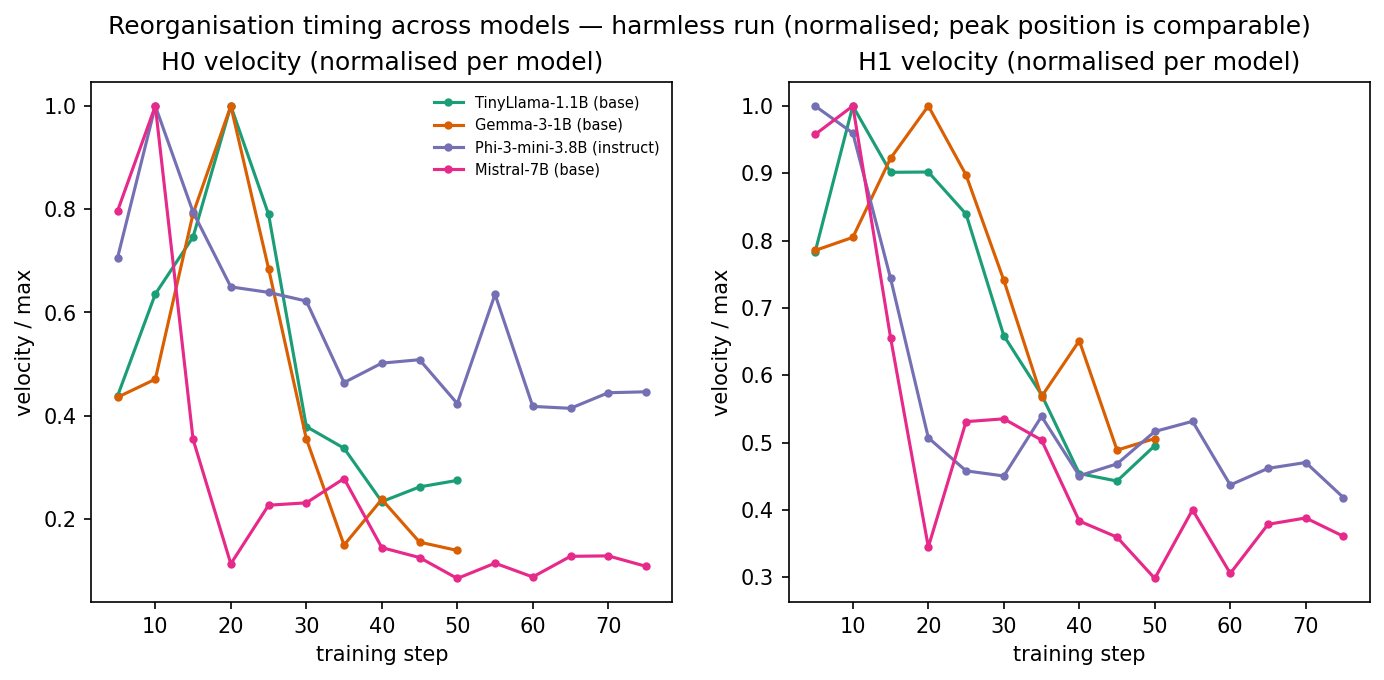}
  \caption{\textbf{The early transient topological peak (RQ1).} Dense early-window $H_0$ (left) and $H_1$ (right)
  Wasserstein velocity, normalized per model so the peak \emph{position} is comparable (magnitudes are scale-dependent). The reorganization is a transient rise-peak-decay whose peak moves earlier with model size.}
  \label{fig:densevel}
\end{figure}

Across all models, the majority of topological reorganization occurs during the earliest stages of fine-tuning. 
A dense checkpoint analysis reveals a transient rise--peak--decay pattern in the $H_1$ Wasserstein velocity, 
followed by rapid stabilization (Fig.~\ref{fig:densevel}). 
The timing of this peak depends on model scale, occurring later in the 1B models and earlier in Phi-3 and Mistral-7B. 
Thus, while an early reorganization phase is present across all models, its precise timing varies with model size.

To assess whether this concentration of topological change reflects genuine temporal structure rather than the total amount of movement, we compare the observed trajectories against a checkpoint-order permutation null. 
The early concentration statistic exceeds the null expectation in 11 of 12 model--objective combinations 
and remains significant in all but one case. Moreover, the same rise--peak--decay pattern is reproduced 
across independent fine-tuning seeds for TinyLlama, Gemma, and Mistral (Appendix \ref{app:seedreorg}), 
indicating that the phenomenon is robust to initialization and data ordering.

For Gemma-3-1B, the $H_1$ velocity rises, peaks near step $20$, and decays along nearly the same curve for \emph{every} seed and objective (Appendix~Fig.~\ref{fig:seeds}, left). 
The early concentration exceeds the checkpoint-order null in all nine seed-objective combinations. 
The early shock is therefore robust across scales, objectives, and seeds.

\subsection{Objectives Leave Distinct, Oppositely-Signed Signatures (RQ2)}
\label{sec:res-sep}

\begin{figure}[t]
  \centering
  \includegraphics[width=0.42\linewidth]{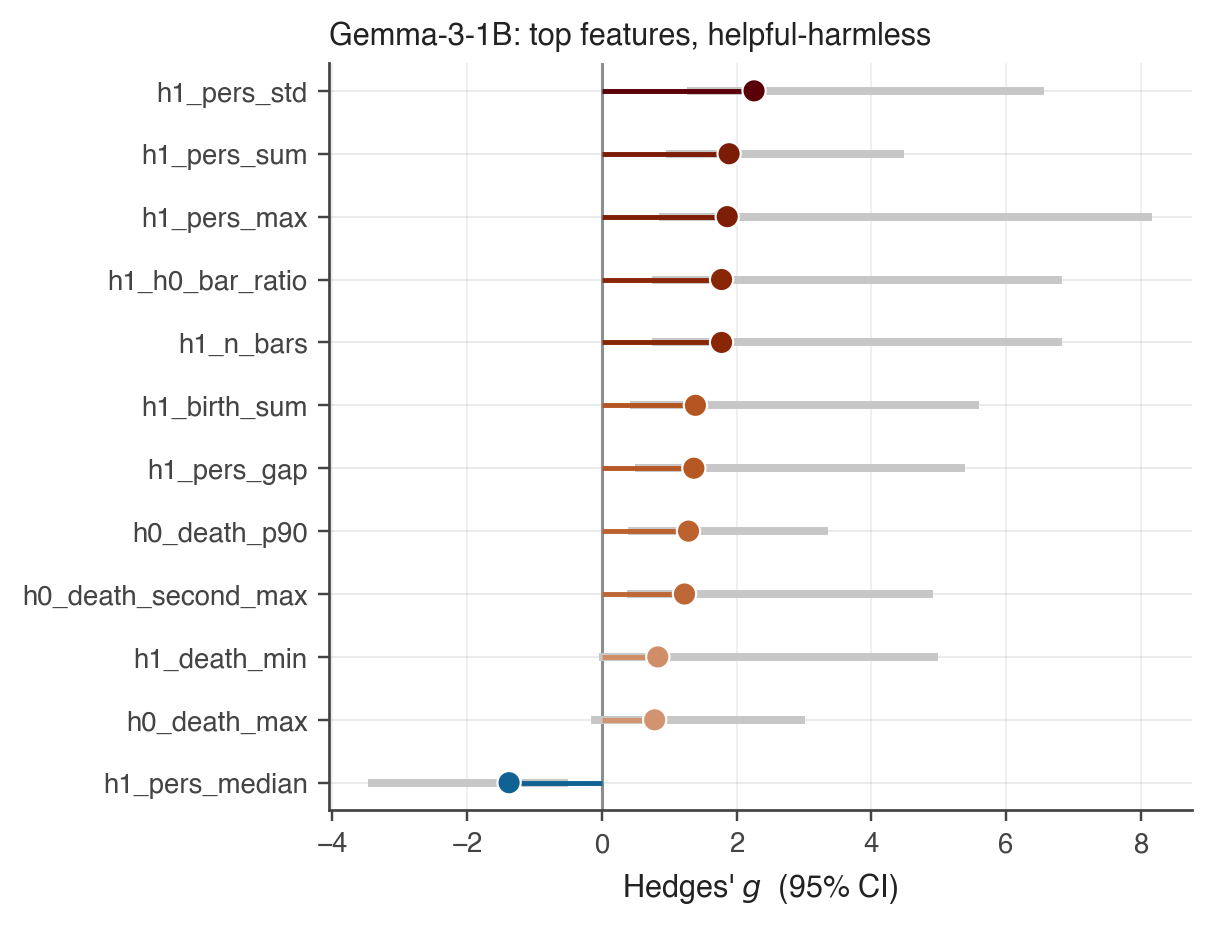}
  \hspace{1em}
  \includegraphics[width=0.48\linewidth]{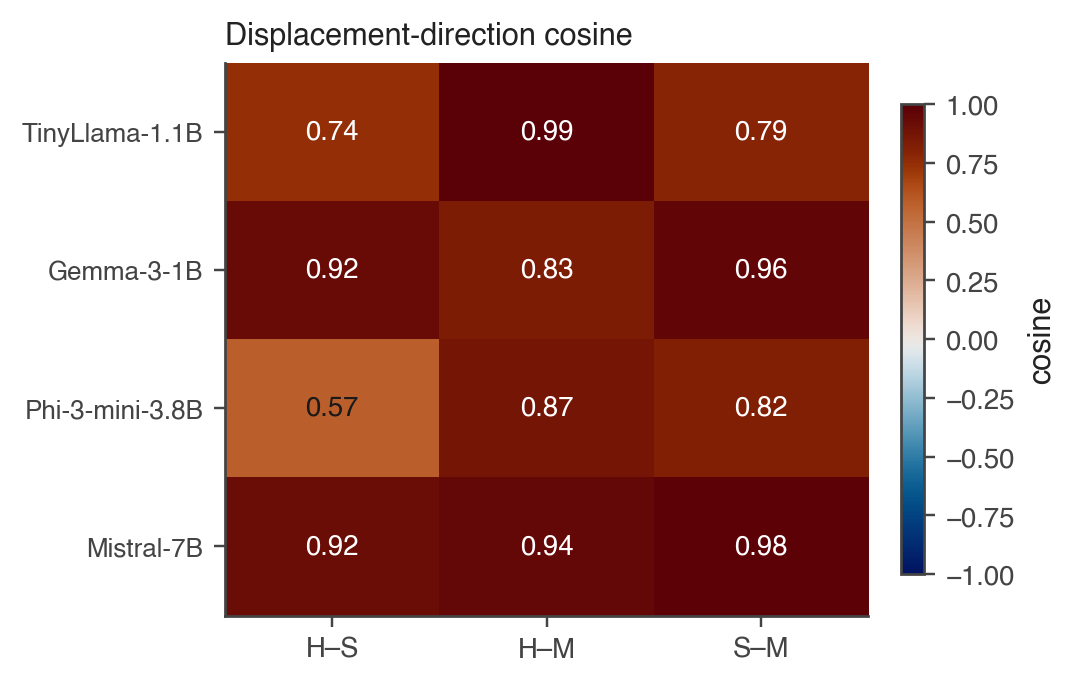}
  \caption{\textbf{Objective Separation (RQ2).} \emph{Left:} Per-feature %cell-level 
  effect sizes (Hedges' $g$, $95\%$ bootstrap CI) for the most-separated pair (Gemma); a few $H_1$-persistence features carry
  $|g|\!\sim\!1$--$2$. \emph{Right:} Objective displacement-direction cosines (model $\times$ pair) are
  all strongly positive; the objectives move the topology the same way, separating in
  magnitude.}
  \label{fig:sep}
\end{figure}

The three alignment objectives induce distinguishable topological trajectories. A small subset of $H_1$-persistence features carries most of the separation signal, with effect sizes reaching Hedges' $|g|\approx1$--$2$ (Fig.~\ref{fig:sep}, left). At the multivariate level, energy distances remain consistently positive across models, indicating persistent separation in the full topological summary vectors. Throughout, we report cell-level effect sizes with bootstrap confidence intervals rather than subsample-level significance tests.

The qualitative direction of topological change depends strongly on the starting state (Fig.~\ref{fig:ordering}, right). In the three base models, fine-tuning reduces $H_0$ persistence entropy relative to the initial checkpoint, with the harmless objective typically producing the strongest compression. In contrast, the instruction-tuned Phi-3 exhibits the opposite behavior, increasing $H_0$ persistence entropy above its initial value. Thus, initialization appears to play a stronger role than model scale in determining whether alignment induces topological compression or enrichment.

Within each model, however, the three objectives evolve along a largely shared axis. Pairwise displacement cosines remain strongly positive (median $0.79$--$0.94$; Fig.~\ref{fig:sep}, right), indicating that helpful, harmless, and mixed fine-tuning differ primarily in the magnitude rather than the direction of their topological reorganization, with harmless fine-tuning generally producing the largest displacement.

Finally, objective separation emerges primarily at the endpoint of training. During the early transient reorganization phase, the three objectives remain nearly indistinguishable ($U=23.6$, $p=0.92$; Appendix~Fig.~\ref{fig:seeds}), suggesting that the initial topological burst is largely shared across objectives. Additional token-pooling experiments indicate that the resulting objective-specific signatures are concentrated near the decision token (Appendix~\ref{app:supp}).

\subsection{Topology Reveals Changes Beyond Coarse Behavioral Metrics (RQ3)}
\label{sec:res-leadlag}

\begin{figure}[t]
  \centering
  \includegraphics[width=0.37\linewidth]{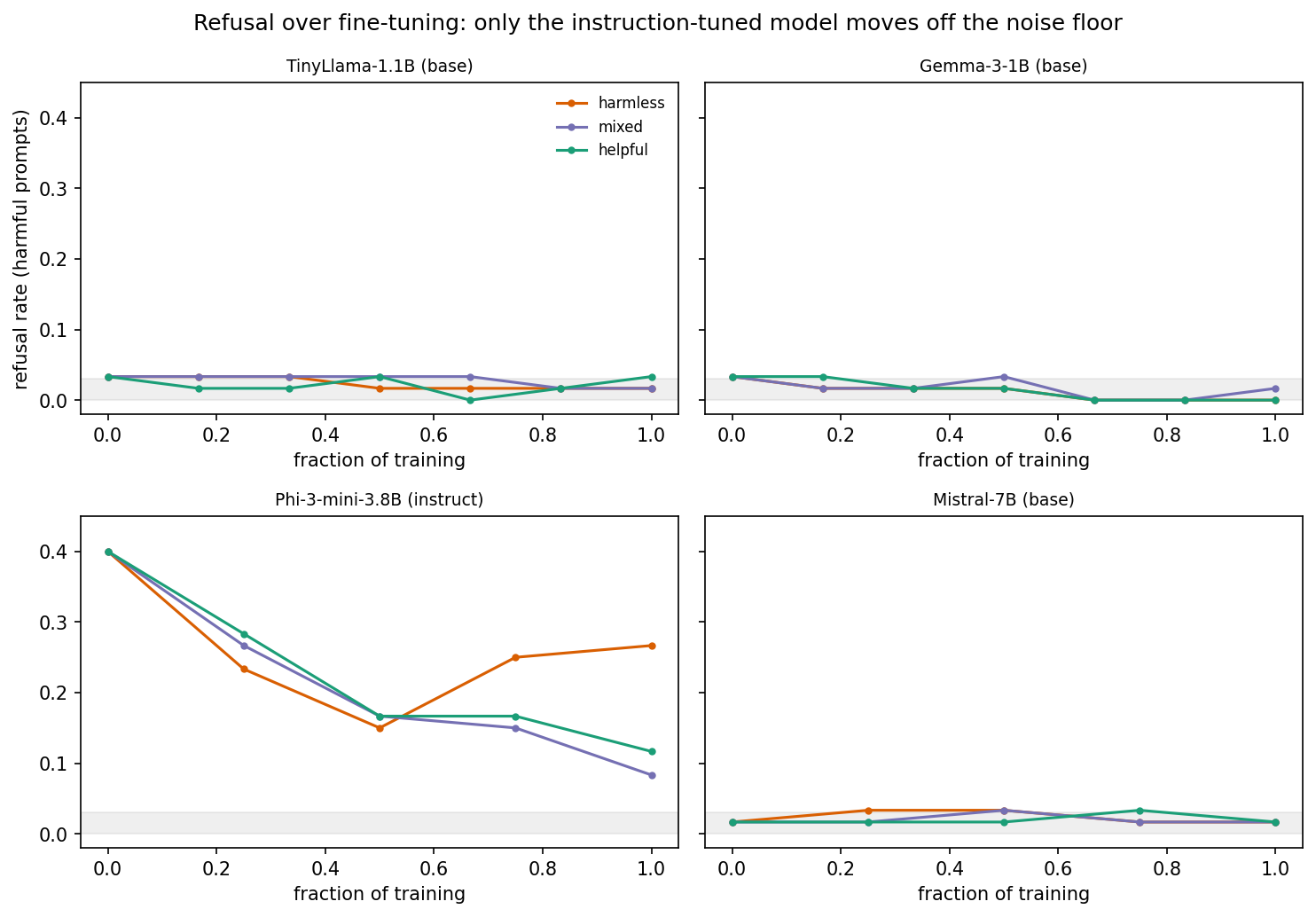}
  \hspace{1em}
  \includegraphics[width=0.49\linewidth]{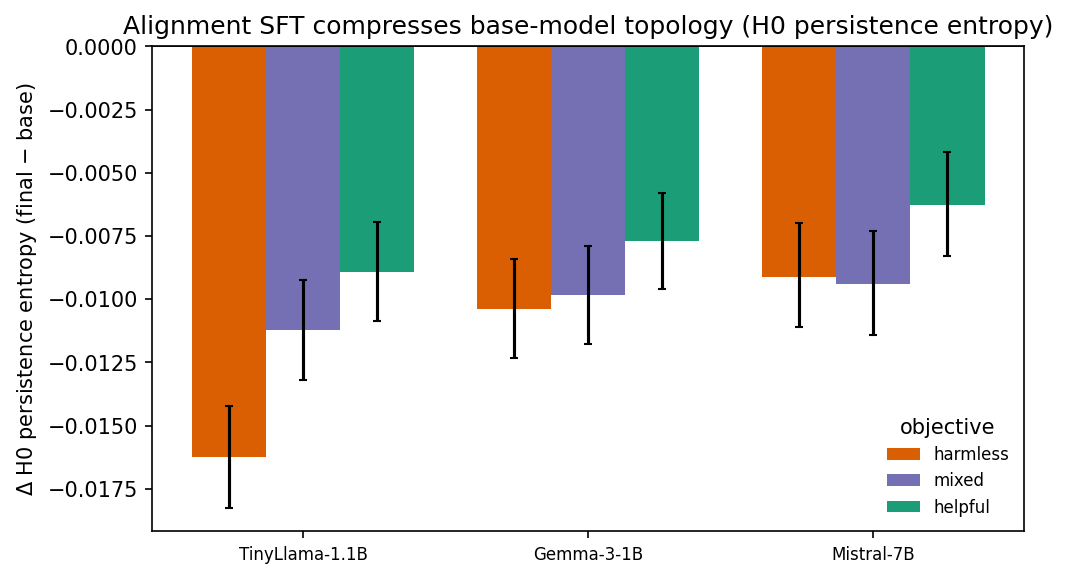}
  \caption{\textbf{Behavior and Objective Ordering (RQ3, RQ2).} \emph{Left:} Refusal on harmful prompts
  over fine-tuning. All three base models stay on the $\approx2$--$3\%$ noise
  floor (shaded); only the instruction-tuned Phi-3 leaves it, reaching $8$--$40\%$, and there the
  objectives \emph{separate}. \emph{Right:} Base-to-final change in $H_0$
  persistence entropy (headline layer, bootstrap $95\%$ CI): base models compress (below zero, harmless
  most), Phi-3 enriches distinct, oppositely-signed signatures set by initialization.}
  \label{fig:ordering}
\end{figure}

Whether alignment becomes behaviorally visible depends strongly on the starting state rather than model scale. Across TinyLlama, Gemma, and Mistral, refusal rates on harmful prompts remain close to their $\approx2$--$3\%$ noise floor throughout fine-tuning and exhibit no clear phase transition (Fig.~\ref{fig:ordering}, left). In contrast, the instruction-tuned Phi-3 undergoes substantial behavioral change, with refusal rates increasing to $8$--$40\%$ and separating by objective. These observations suggest that, under low-rank fine-tuning, pretrained models can undergo substantial representation-level reorganization without corresponding changes in coarse behavioral metrics.

A more sensitive behavioral measure based on Jensen--Shannon divergence between successive next-token distributions confirms that the model's output distribution does evolve during fine-tuning (Appendix~\ref{app:js_velocity}). However, topological velocity consistently peaks no later than behavioral velocity and in several cases slightly earlier (Appendix~Fig.~\ref{fig:leadlag}; Table~\ref{tab:appleadlag}). Together, these findings suggest that persistent homology captures representation-level changes that are only partially reflected in observable behavior.

\subsection{Robustness and Geometric Controls}
\label{sec:res-floor}

Several control analyses indicate that the observed topological phenomena do not reduce to simple geometric statistics. Relative to isotropic Gaussian point clouds matched in dimension and sample size, the informative signal does not reside in individual scalar summaries such as raw $H_0$ entropy alone, but rather in the multivariate evolution of the topological descriptors.

Moreover, although centroid drift, total variance, and mean pairwise distance all detect the initial transient reorganization, these geometric statistics return rapidly to baseline, whereas the $H_1$ Wasserstein velocity remains elevated (Appendix~Figs.~\ref{fig:geomablation},~\ref{fig:appgeomphi3}). PH therefore tracks ongoing representation reorganization that is not captured by changes in the cloud's centroid, scale, or overall spread. Additional controls, including checkpoint-order nulls, token-level pooling analyses, and subsample robustness checks, are reported in Appendix~\ref{app:controls}.

\section{Conclusion and Future Directions}
\label{sec:discussion}

Across all models examined, alignment fine-tuning induces an early transient phase of topological reorganization whose timing varies modestly with model scale. Different alignment objectives subsequently produce distinguishable trajectories, and the direction of these changes depends strongly on initialization.
PH reveals representation-level changes that are only partially reflected in coarse behavioral metrics and therefore provides a useful lens for studying the dynamics of internal representations during alignment. An interesting direction for future work is to investigate whether similar topological phenomena arise in multimodal foundation models.

\section*{Acknowledgments}

K.F.~and A.M.~are supported by the EPSRC AI Hub on Mathematical Foundations of Intelligence: An ``Erlangen Programme'' for AI [EP/Y028872/1].

\clearpage
\bibliographystyle{authordate3}
\bibliography{refs}

\clearpage
\appendix

\section{Supplementary Results}
\label{app:supp}

\subsection{Robustness Checks and Controls}
\label{app:controls}

We performed a series of controls to verify that the observed topological
phenomena are not artifacts of checkpoint ordering, simple geometric
statistics, token selection, or the use of overlapping subsamples.

\begin{figure}[p]
  \centering
  \includegraphics[width=.8\linewidth]{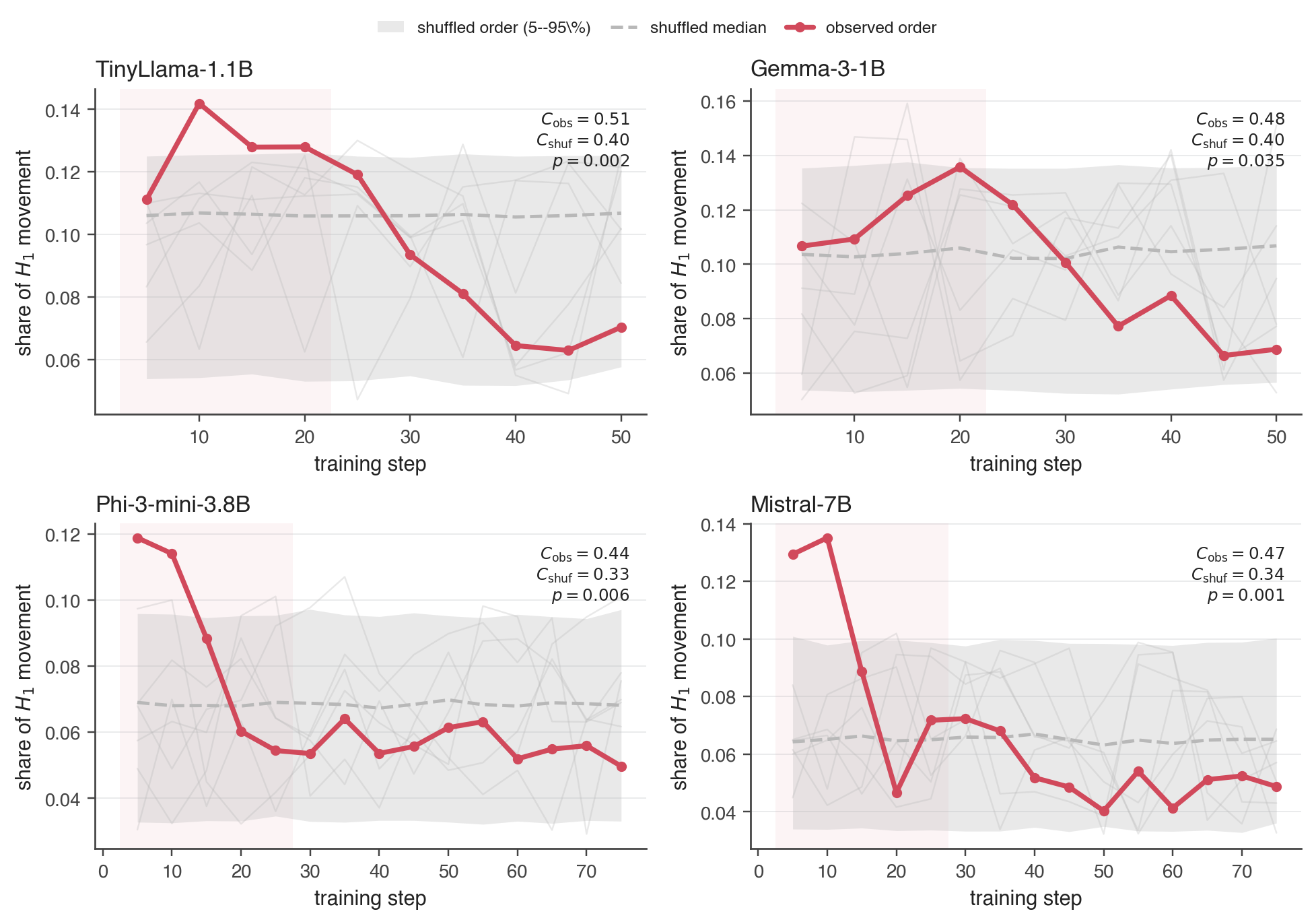}
  \caption{\textbf{Permuting the checkpoint order removes the early peak.} 
  For each dense model (harmless run), the share of total $H_1$ Wasserstein movement at each step under the \emph{observed} checkpoint order (red, concentrated in the shaded first-third window) versus $1000$ random checkpoint orders (grey median, $5$--$95\%$ band, and a few individual traces).}
  % Time-shuffled orders spread the movement roughly uniformly, so the observed early concentration $C_{\mathrm{obs}}$ exceeds the shuffled mean
  % $C_{\mathrm{shuf}}$ with the permutation $p$ shown ($11/12$ cells significant across all objectives;
  % cf.\ Table~\ref{tab:appshock}).}
  \label{fig:shufflecurves}
\end{figure}

\begin{figure}[p]
  \centering
  \includegraphics[width=.6\linewidth]{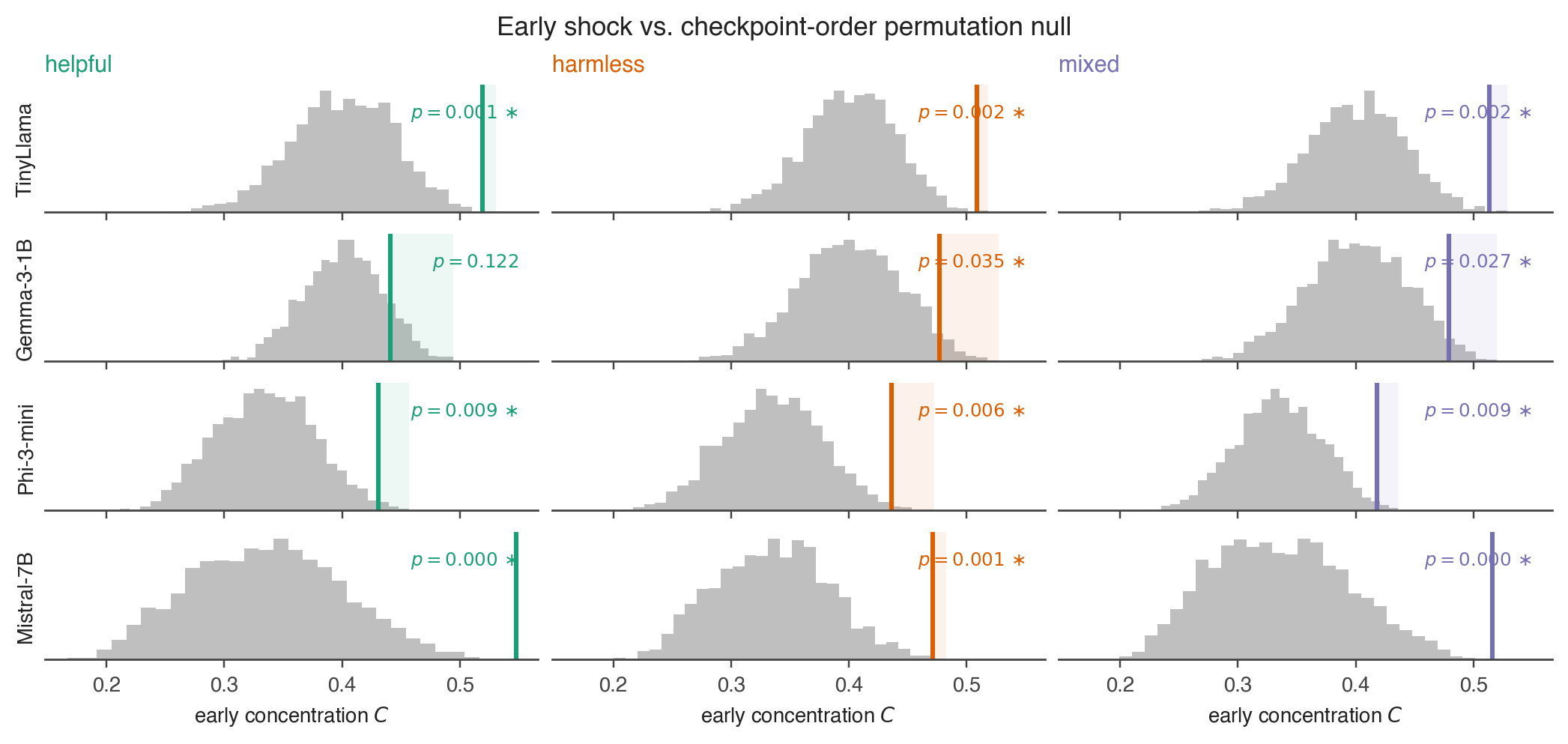}
  \caption{\textbf{The early shock against a checkpoint-order permutation null} (\S\ref{sec:res-shock}). For each (model, objective), the null distribution of the early-velocity concentration $C$ ($H_1$ Wasserstein) under randomly shuffled checkpoint order (grey, $2000$ permutations), with the observed $C$ as a colored line (shaded tail) and its permutation $p$; $\ast\,p{<}0.05$.}
  % The observed concentration lies in the upper tail in $11/12$ cells. This means that the burst is genuinely early and not an artifact of total movement.}
  \label{fig:shufflenull}
\end{figure}

\paragraph{Checkpoint-order null.}
Permuting checkpoint order and recomputing the $H_1$
Wasserstein velocity destroys the early concentration of
topological movement.
Under the observed checkpoint ordering, the velocity is strongly
concentrated within the first third of training, whereas random
checkpoint permutations distribute the same total movement
approximately uniformly across time
(Figs.~\ref{fig:shufflecurves}--\ref{fig:shufflenull},
Table~\ref{tab:appshock}).

\begin{table}[p]
  \centering\small
  \begin{tabular}{lccl}
    \toprule
    Model & Significant ($p{<}0.05$) & $p$ range & Peak position \\
    \midrule
    TinyLlama-1.1B  & $3/3$ & $0.002$--$0.003$ & first third \\
    Gemma-3-1B      & $2/3$ & $0.024$--$0.12$  & first third \\
    Phi-3-mini-3.8B & $3/3$ & $0.008$--$0.010$ & earliest \\
    Mistral-7B      & $3/3$ & $\approx 0.001$  & earliest \\
    \bottomrule
  \end{tabular}
  \caption{Early-shock checkpoint-order null per model across the three objectives 
  ($H_1$ Wasserstein velocity). The early velocity concentration beats the time-shuffled null in $11/12$ cells. 
  See \texttt{results/metrics/exp2\_cell/shuffle\_null.csv} for
  the full per-cell values.}
  \label{tab:appshock}
\end{table}

\begin{table}[p]
  \centering\small
  \begin{tabular}{llccl}
    \toprule
    Model & Start & $U$ (standardized) & Single-run $U$-test & Energy dist.\ range \\
    \midrule
    TinyLlama-1.1B  & base     & $2.33$  & degenerate ($p$ undef.) & $0.31$--$0.55$ \\
    Gemma-3-1B      & base     & $73.7$  & degenerate ($p$ undef.) & $1.02$--$1.35$ \\
    Phi-3-mini-3.8B & instruct & $12.0$  & degenerate ($p$ undef.) & --- \\
    Mistral-7B      & base     & $0.44$  & degenerate ($p$ undef.) & --- \\
    \bottomrule
  \end{tabular}
  \caption{Cell-level objective separation. The all-pairs $U$ is invariant to relabeling with one  trajectory per objective, so the single-run permutation test is degenerate for every model.}
  % the structural reason the seed-replicated test of \S\ref{sec:res-seeds} is needed.}
  \label{tab:appU}
\end{table}

\begin{table}[p]
  \centering\small
  \begin{tabular}{llccc}
    \toprule
    Model & Objective & Topo.\ peak & Func.\ peak & Lead (steps) \\
    \midrule
    Gemma-3-1B  & helpful  & $15$ & $20$ & $+5$ \\
    Gemma-3-1B  & harmless & $20$ & $20$ & $0$ \\
    Gemma-3-1B  & mixed    & $20$ & $20$ & $0$ \\
    Phi-3-mini  & helpful  & $10$ & $15$ & $+5$ \\
    Phi-3-mini  & harmless & $5$  & $10$ & $+5$ \\
    Phi-3-mini  & mixed    & $5$  & $5$  & $0$ \\
    Mistral-7B  & helpful  & $5$  & $10$ & $+5$ \\
    Mistral-7B  & harmless & $10$ & $10$ & $0$ \\
    Mistral-7B  & mixed    & $5$  & $10$ & $+5$ \\
    \bottomrule
  \end{tabular}
  \caption{Velocity-peak lead-lag (\S\ref{sec:res-leadlag}). Lead $=$ function-peak $-$ topology-peak
  step. Topology never lags, leading by at most one checkpoint. Source:
  \texttt{results/metrics/exp2\_cell/leadlag.csv}.}
  \label{tab:appleadlag}
\end{table}

\begin{figure}[ptb]
  \centering
  \includegraphics[width=.75\linewidth]{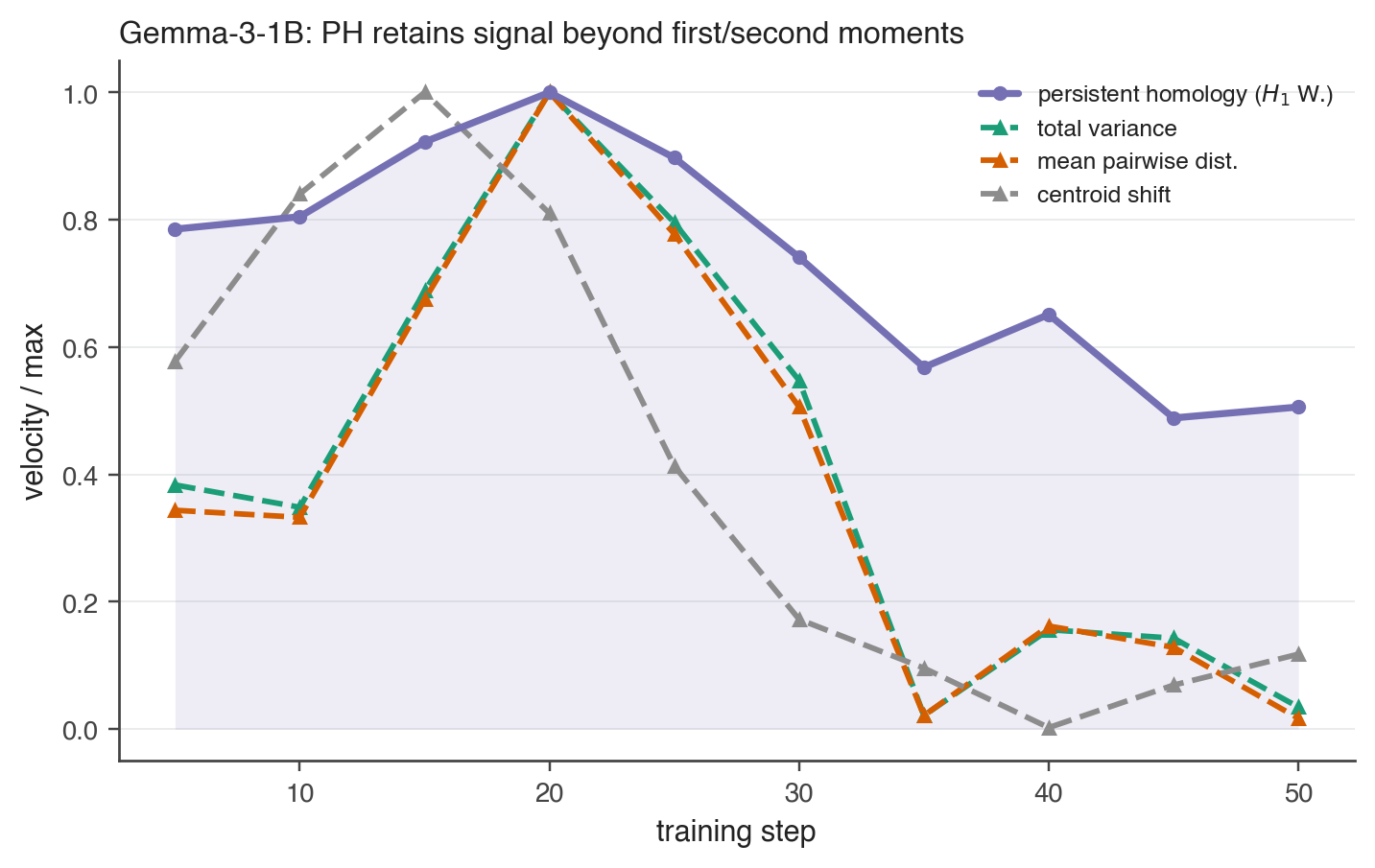}
  \caption{\textbf{PH vs.\ geometry (Gemma, harmless, dense).} Normalized velocity of the $H_1$ Wasserstein metric against non-topological baselines on the same clouds (\S\ref{sec:res-floor}). All rise in the early burst, but centroid drift and variance collapse to floor by step $\sim\!35$ while PH
  stays elevated; PH is therefore not redundant with first/second moments.}
  \label{fig:geomablation}
\end{figure}

\begin{figure}[p]
  \centering
  \includegraphics[width=.75\linewidth]{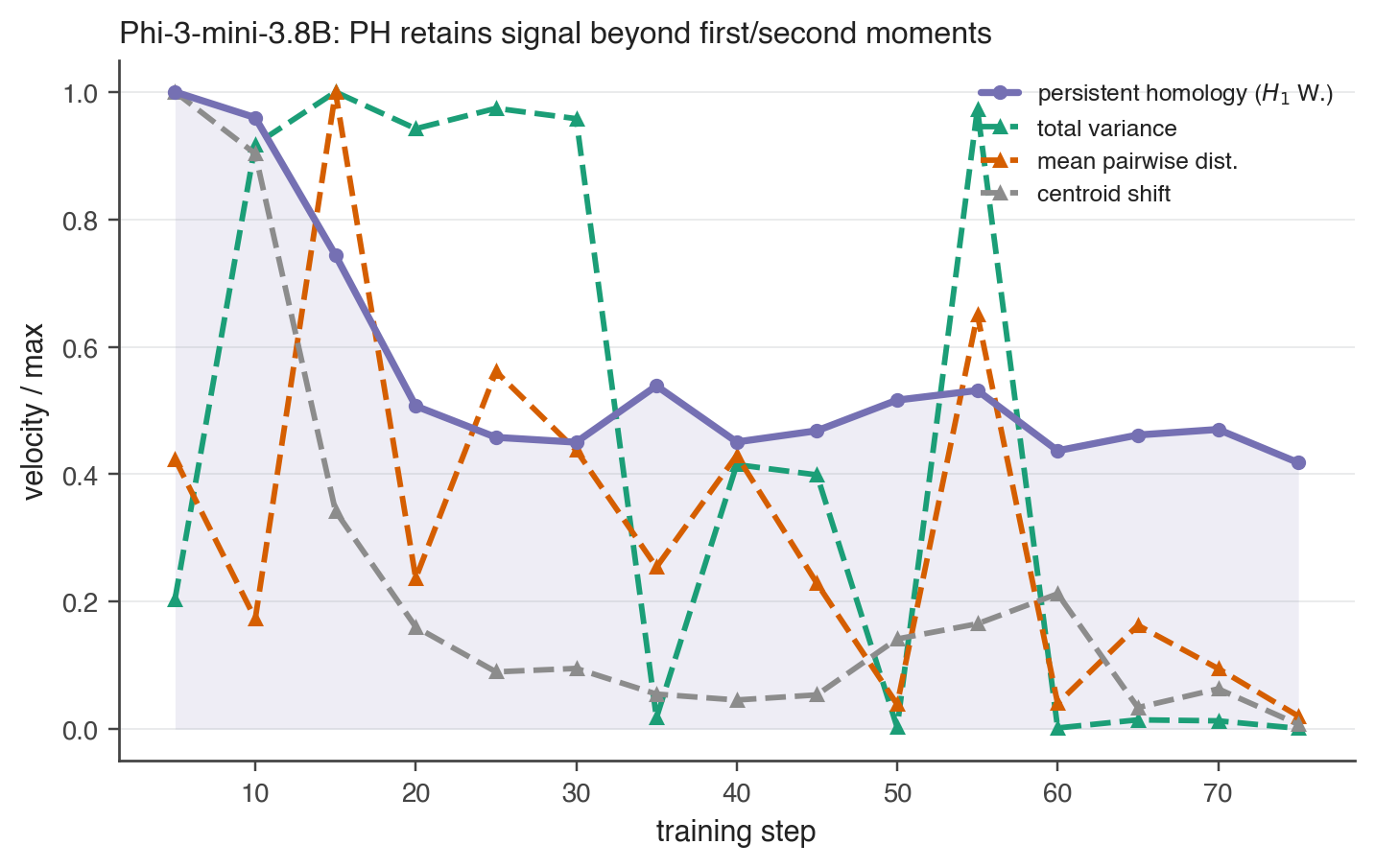}
  \caption{Phi-3 geometric-baseline ablation (companion to Fig.~\ref{fig:geomablation}): the $H_1$
  Wasserstein velocity retains a longer-lived signal than centroid drift / variance / mean pairwise distance, which collapse after the early burst.}
  \label{fig:appgeomphi3}
\end{figure}

\paragraph{Geometric baselines.}
We compare persistent homology with conventional geometric
statistics computed on the identical activation clouds, namely
centroid drift, total variance, and mean pairwise distance.
All measures exhibit the initial transient burst.
However, the geometric statistics rapidly return to baseline,
whereas the $H_1$ Wasserstein velocity remains elevated
(Figs.~\ref{fig:geomablation} and \ref{fig:appgeomphi3}),
indicating that persistent homology is not redundant with
first- and second-order geometric moments.

\paragraph{Isotropic-Gaussian floor.}
An isotropic Gaussian cloud matched in sample size and ambient
dimension produces approximately $238$ artifactual $H_1$ bars and an
$H_0$ persistence entropy of approximately $5.07$, close to the scalar
entropy values observed after fine-tuning.
This motivates the multivariate topological summaries used in
\S\ref{sec:res-floor}.

\paragraph{Token-level pooling.}
Pooling the final $16$ token representations substantially weakens
objective separation (Fig.~\ref{fig:appholm}),
localizing the discriminative structure to the decision token.

\paragraph{Subsample pseudoreplication.}
Treating the $64$ overlapping subsamples of an activation cloud as
independent observations produces highly inflated significance levels
(Fig.~\ref{fig:appdebias}).
Consequently, all confirmatory inference in this work is performed at
the cell level and reported using effect sizes.

\subsection{Objective Separation}
\label{app:cellsep}

Table~\ref{tab:appU} reports the standardized objective-separation
statistic $U$ together with the associated single-run degeneracy.
Because each objective contributes only a single trajectory,
permuting objective labels leaves the pairwise trajectory dispersion
invariant and therefore yields a degenerate permutation test.
This motivates the seed-replicated analysis of
\S\ref{app:seedreorg}.

\begin{figure}[p]
  \centering
  \includegraphics[width=0.62\linewidth]{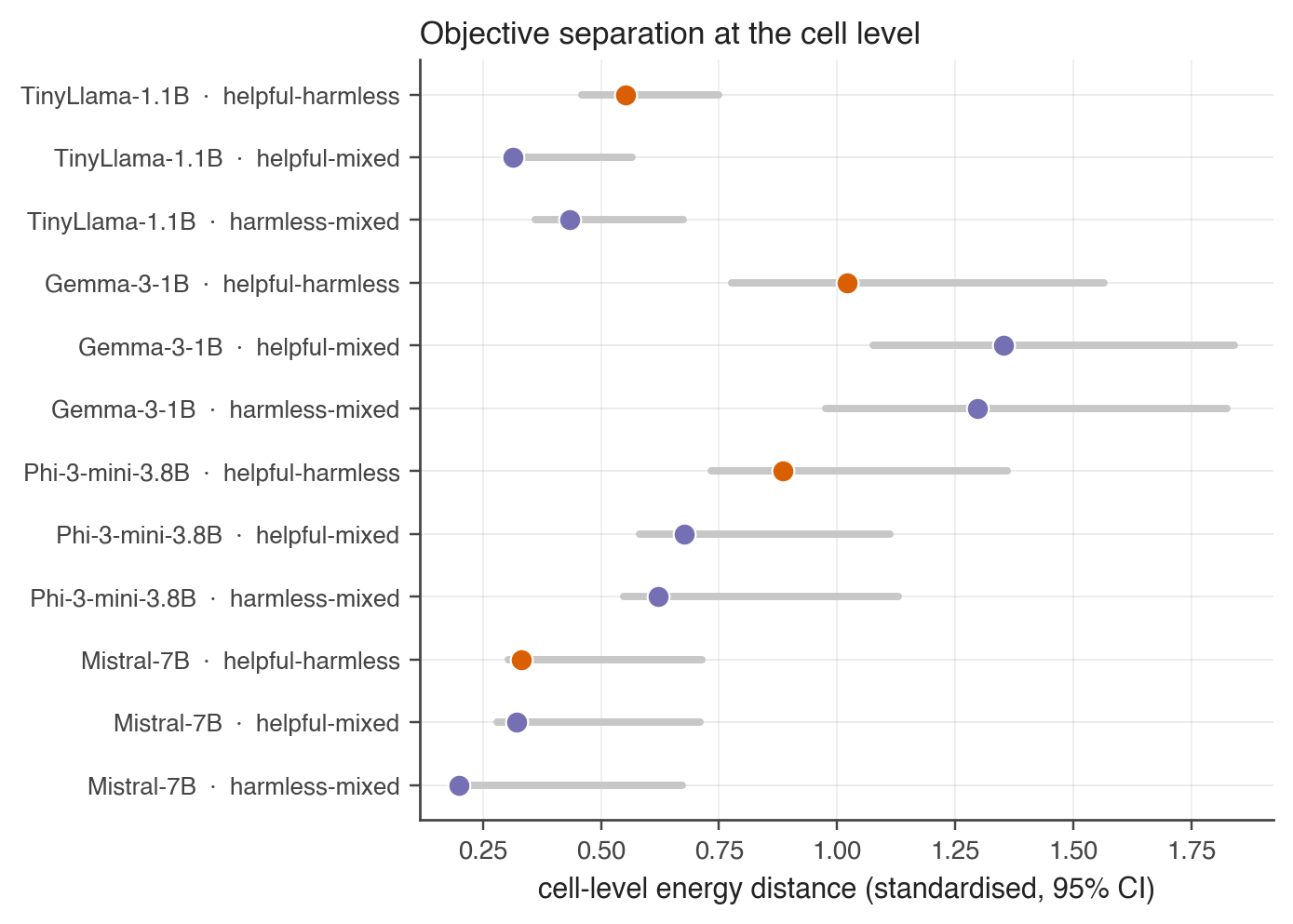}
  \caption{Cell-level energy distance between objectives per pair, per model (standardized, $95\%$
  bootstrap CI). Cell-level energy distances are modest and accompanied by wide intervals, illustrating the more conservative, de-biased view of objective separation.}
  \label{fig:appenergy}
\end{figure}

\begin{figure}[p]
  \centering
  \includegraphics[width=0.62\linewidth]{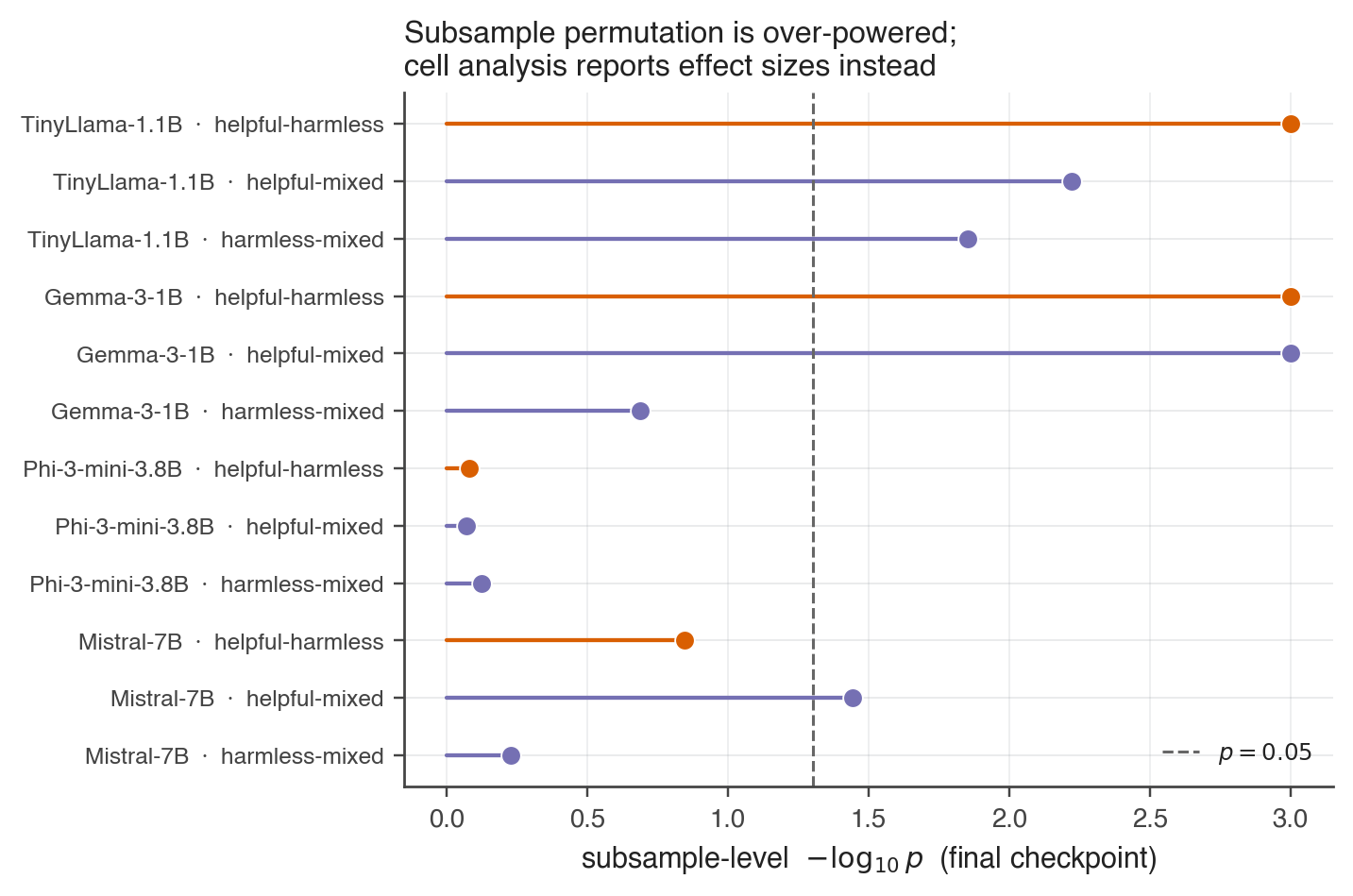}
  \caption{Pseudoreplication inflates significance. Subsample-level permutation $-\log_{10}p$ at the final
  checkpoint, per (model, pair). Treating $64$ overlapping subsamples as independent drives $p$ far below
  $0.05$; the cell-level analysis reports effect sizes (Fig.~\ref{fig:appenergy}, Fig.~\ref{fig:sep})
  instead.}
  \label{fig:appdebias}
\end{figure}

\begin{figure}[ptb]
  \centering
  \includegraphics[width=\linewidth]{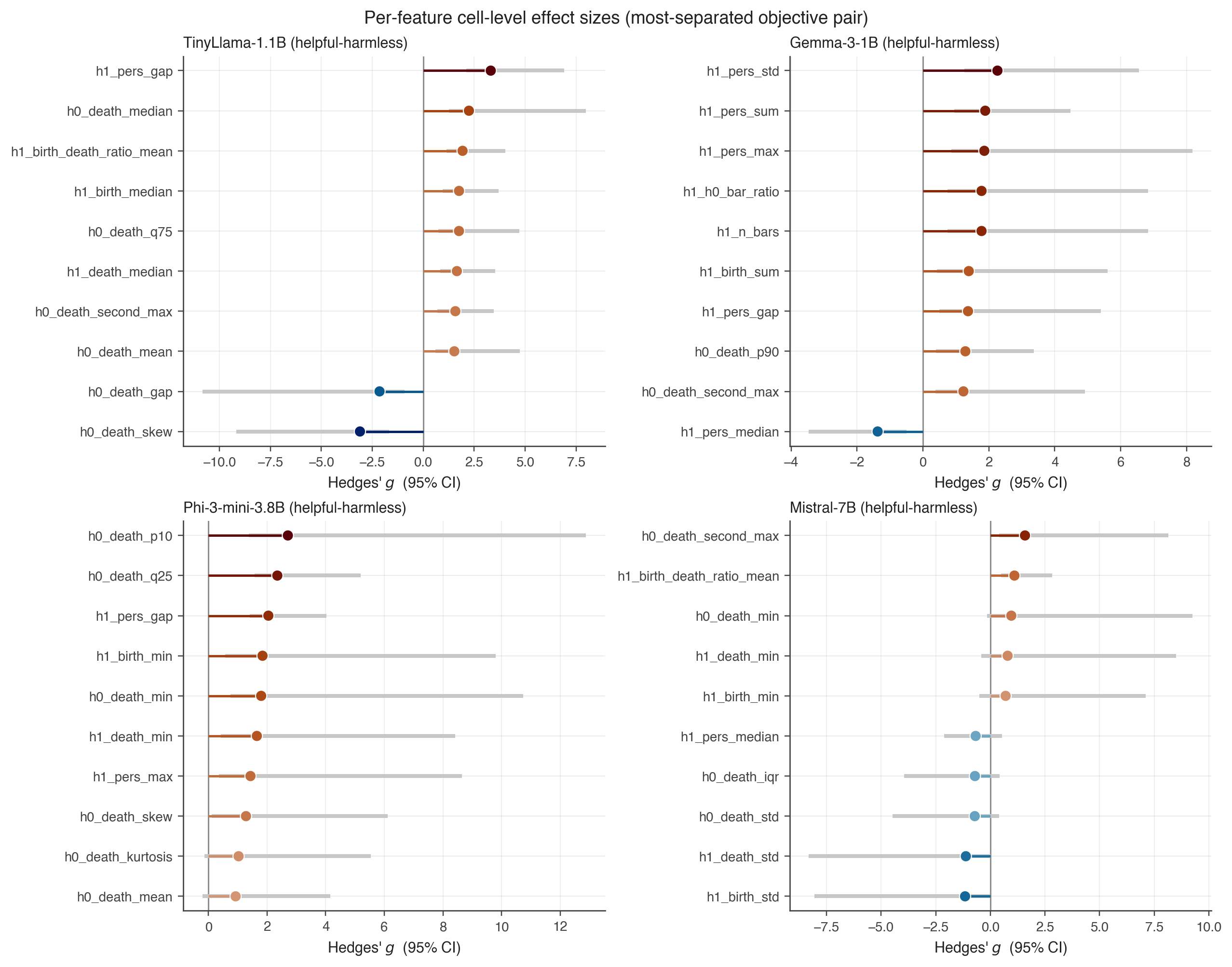}
  \caption{Per-feature cell-level effect sizes (Hedges' $g$, $95\%$ CI) 
           for the most-separated objective pair of each model.
           This is the full version of Fig.~\ref{fig:sep} (left).}
  \label{fig:appforestgrid}
\end{figure}

\begin{figure}[ptb]
  \centering
  \includegraphics[width=\linewidth]{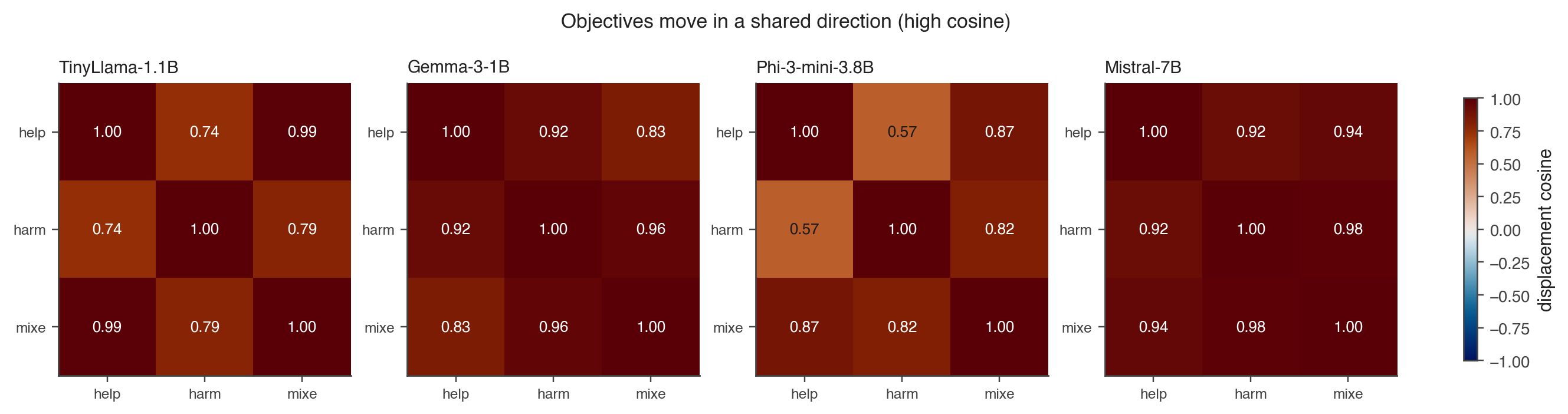}
  \caption{Per-model objective displacement-direction cosine matrices; 
           every off-diagonal entry is strongly positive.
           This is a companion to Fig.~\ref{fig:sep}, right.}
  \label{fig:appcos4}
\end{figure}

Cell-level energy distances are shown in
Fig.~\ref{fig:appenergy}, while per-feature effect sizes and
objective displacement-direction cosines are shown in
Figs.~\ref{fig:appforestgrid} and \ref{fig:appcos4},
respectively.

\subsection{Early Reorganization Across Models}
\label{app:reorg}

\begin{figure}[ptb]
  \centering
  \includegraphics[width=\linewidth]{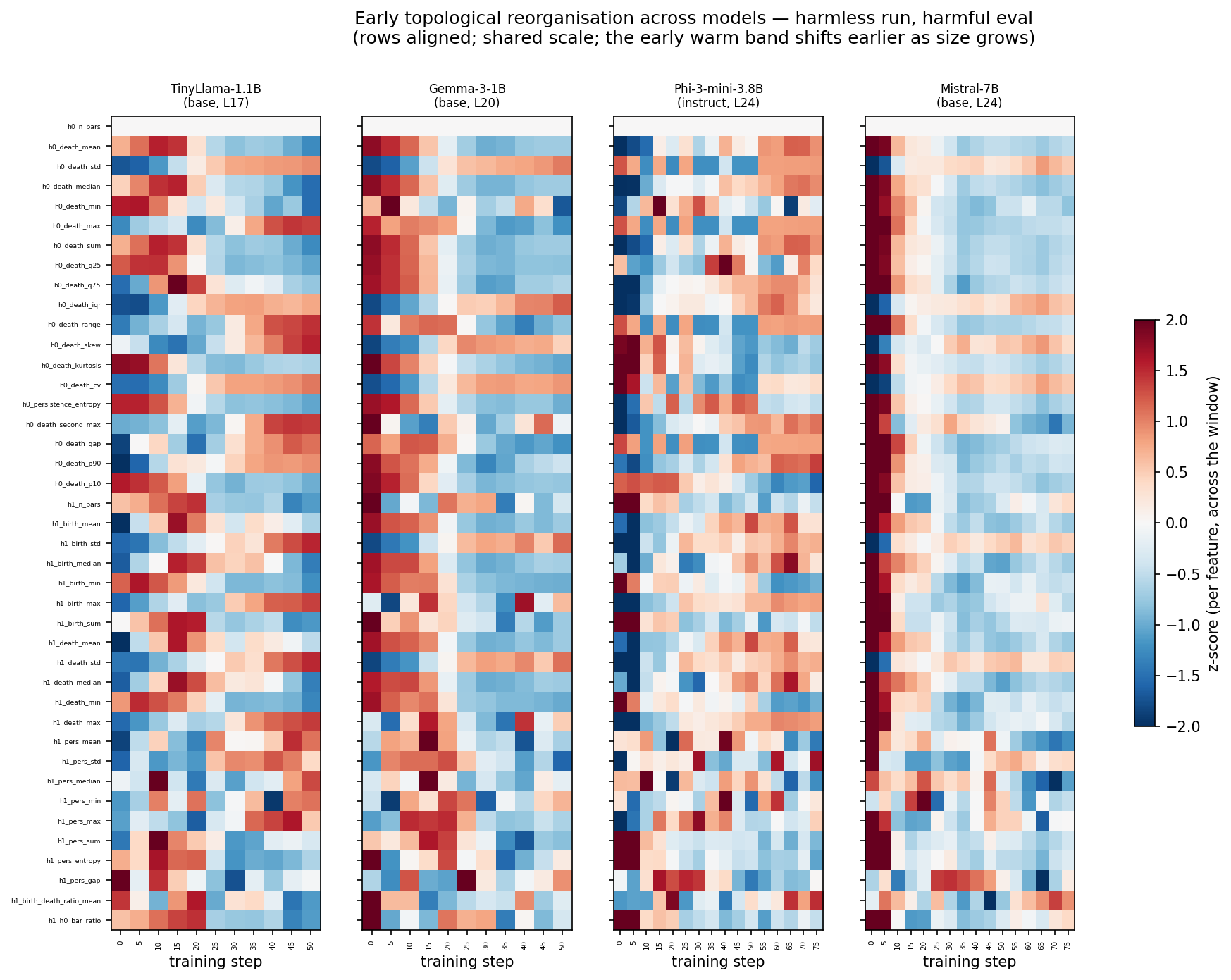}
  \caption{Per-feature reorganisation across the four models (harmless run, harmful eval), each column
  $z$-scored across its own window with one shared diverging scale and identical $41$-feature row order.}
  \label{fig:appreorggrid}
\end{figure}

Figure~\ref{fig:appreorggrid} shows the
$z$-scored feature-by-step reorganization maps for all four models.
In every case, the dominant reorganization is concentrated in an
early high-activity band.
As model scale increases, this band shifts progressively earlier in
training.

\begin{figure}[ptb]
  \centering
  \includegraphics[width=0.6\linewidth]{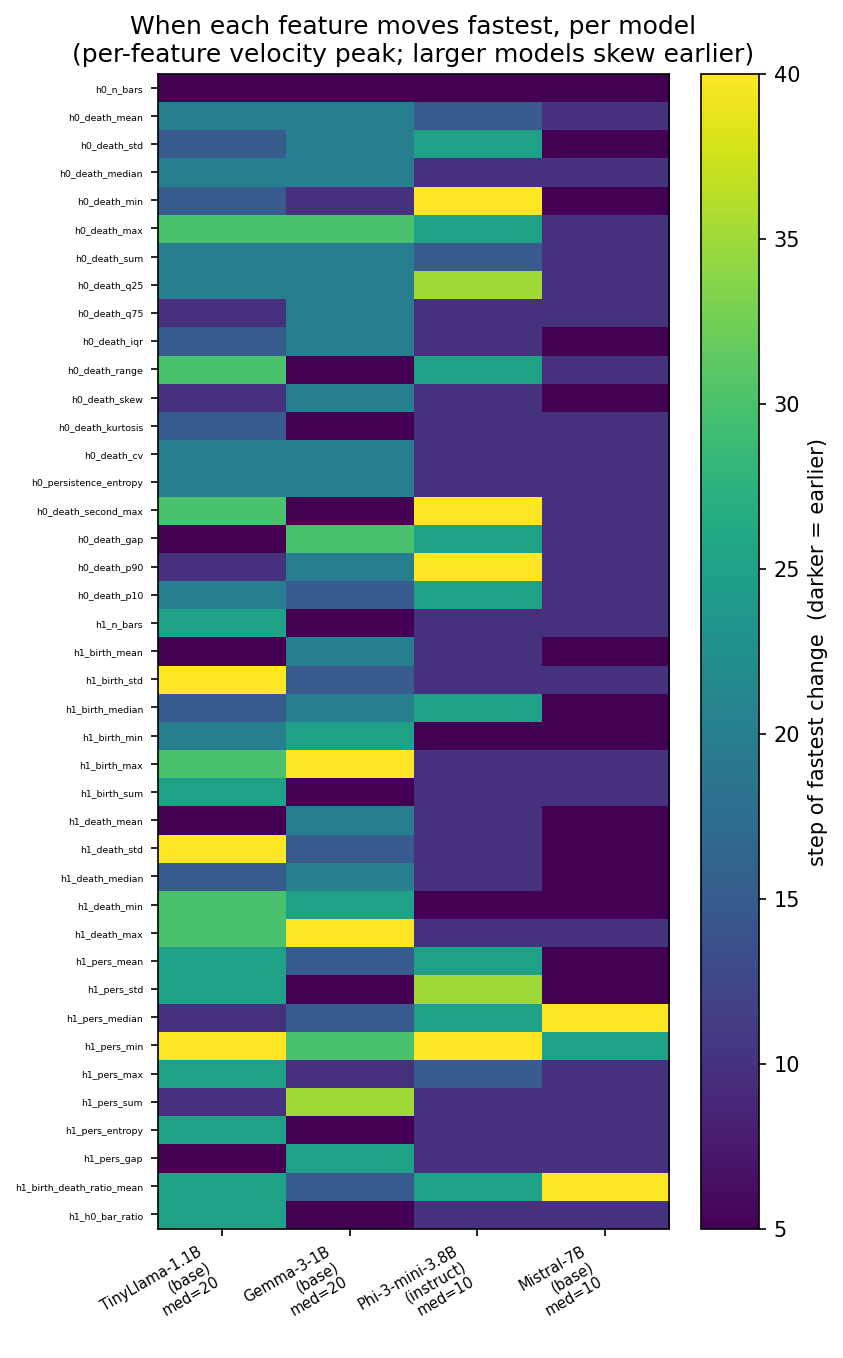}
  \caption{Dense early-window per-feature peak step ($41$ features $\times$ $4$ models): larger models
  skew earlier, the per-feature companion to the velocity burst of Fig.~\ref{fig:densevel}.}
  \label{fig:apppeak}
\end{figure}

\begin{figure}[ptb]
  \centering
  \includegraphics[width=0.49\linewidth]{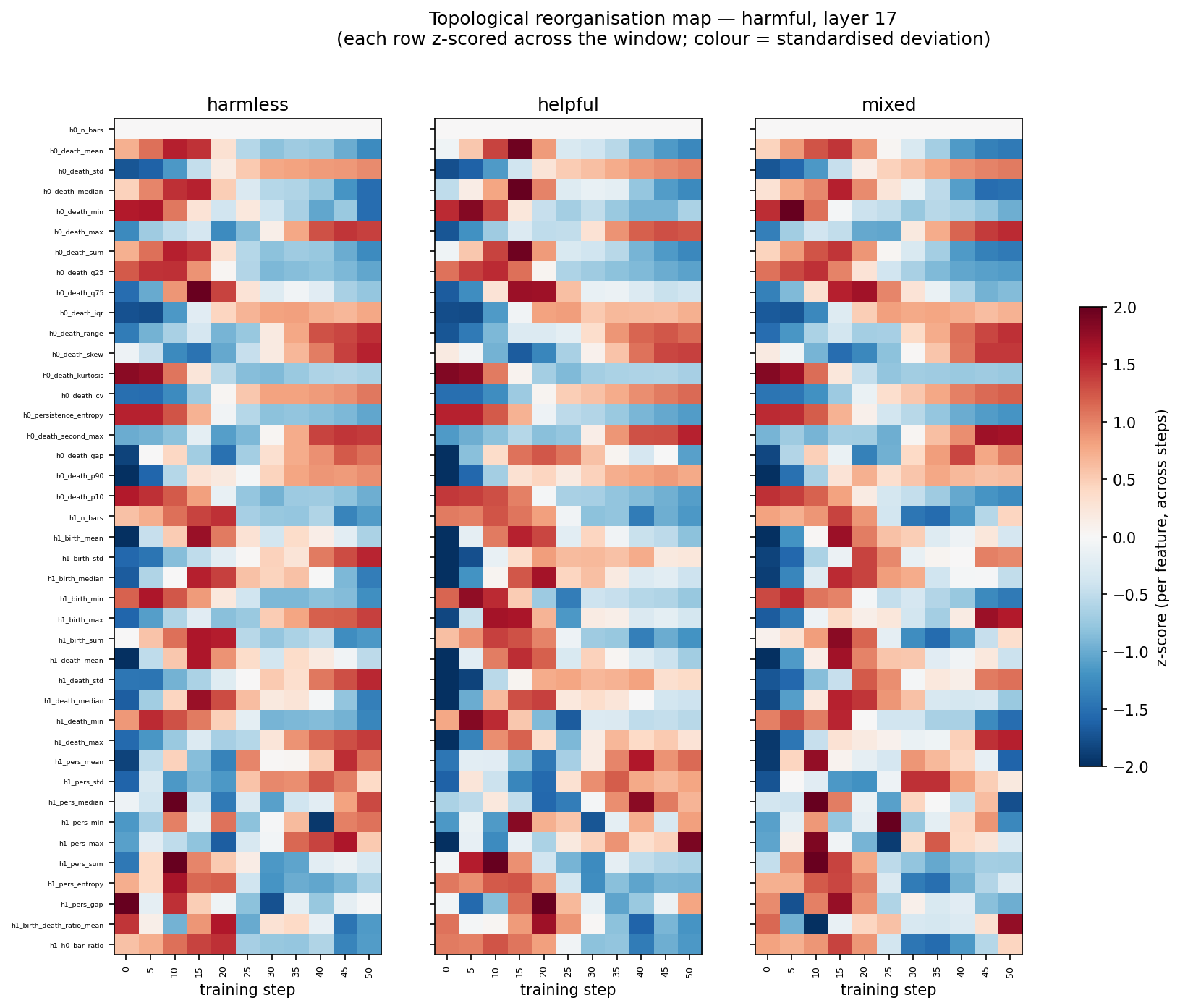}\hfill
  \includegraphics[width=0.49\linewidth]{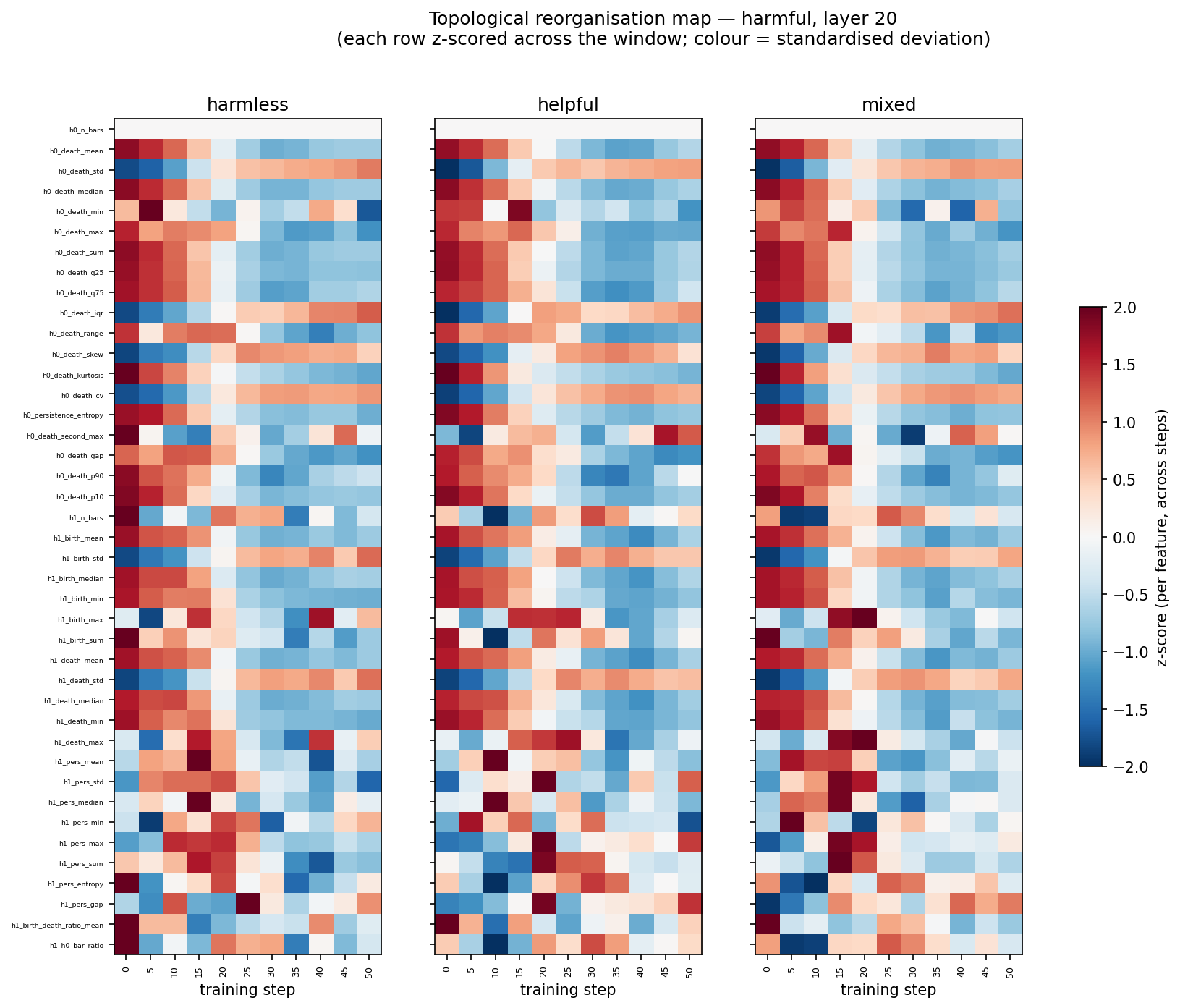}\\[3pt]
  \includegraphics[width=0.49\linewidth]{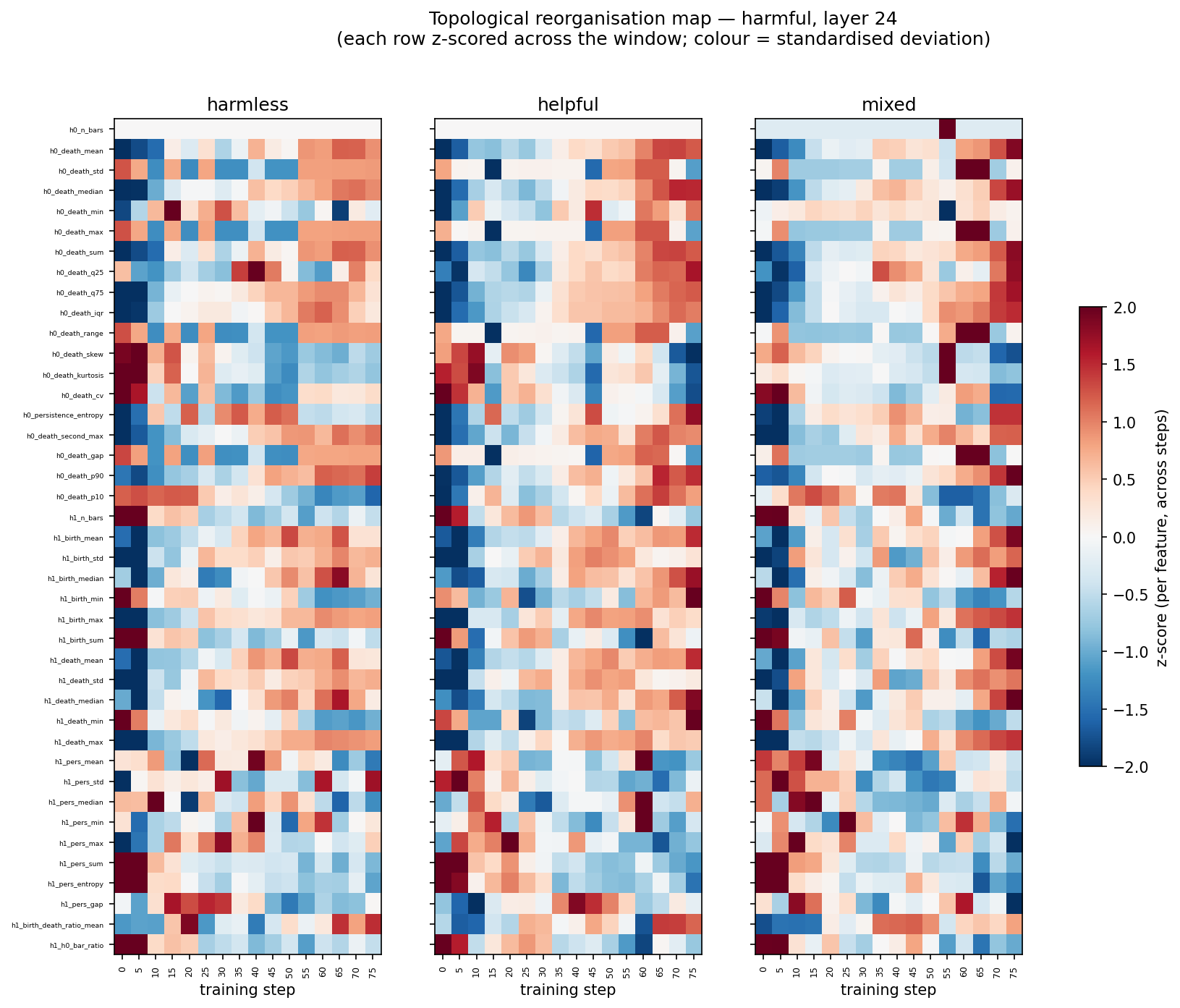}\hfill
  \includegraphics[width=0.49\linewidth]{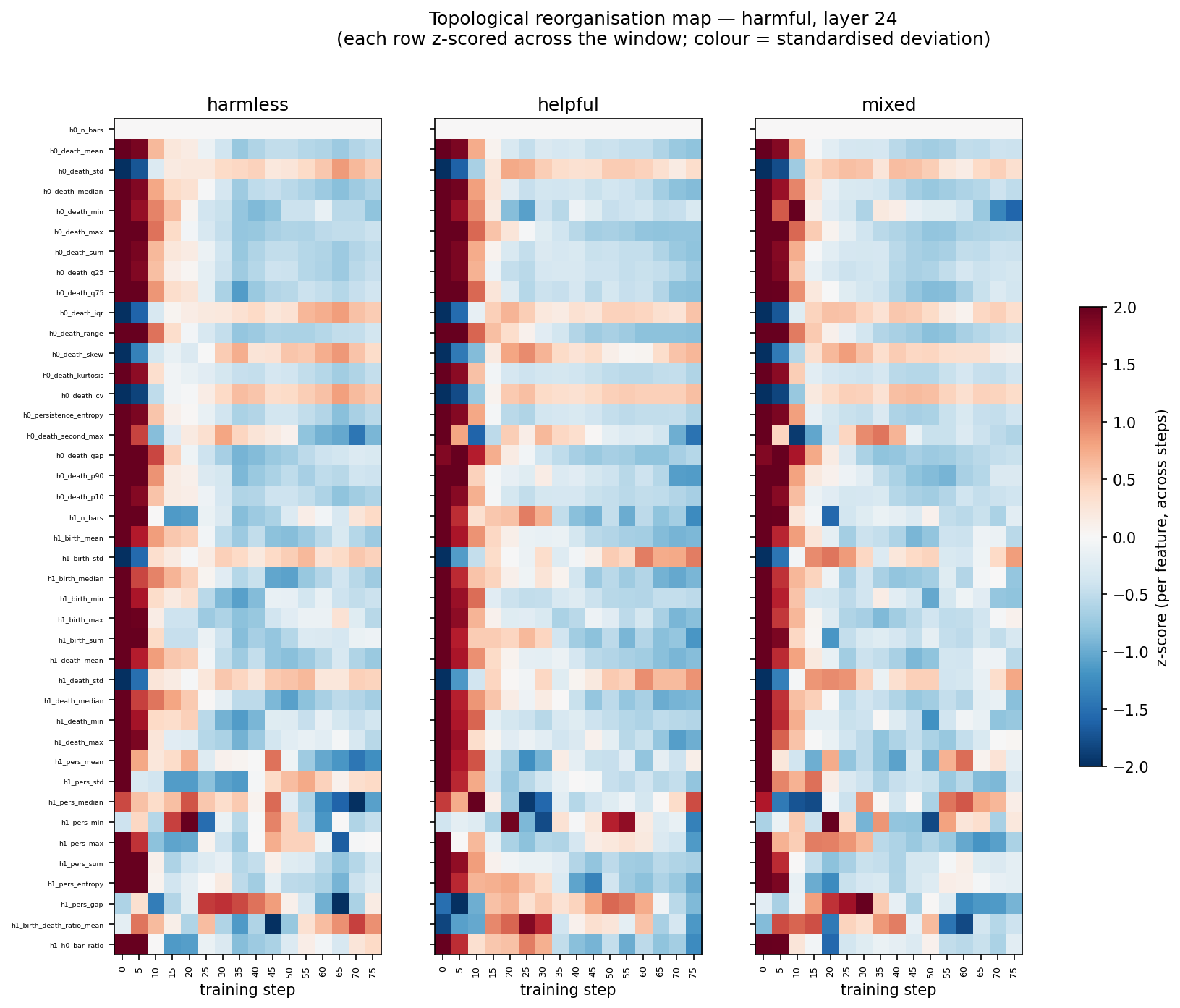}
  \caption{Per-model dense reorganisation maps (harmless run): TinyLlama-1.1B (top left), Gemma-3-1B
  (top right), Phi-3-mini-3.8B (bottom left), Mistral-7B (bottom right). The early high-activity band
  concentrates and moves earlier as model size grows.}
  \label{fig:appreorgheat}
\end{figure}

Figure~\ref{fig:apppeak} summarizes the peak step of each of the
$41$ barcode features across models.
Figure~\ref{fig:appreorgheat} shows the corresponding
per-model heatmaps at their native resolutions.
The associated dense-window velocity trajectories are given in
Fig.~\ref{fig:densevel} of the main text.

\subsection{Seed Replication}
\label{app:seedreorg}

To assess robustness, we replicated the dense early-window
experiments using three independent seeds, varying both LoRA
initialization and data order.

\begin{figure}[p]
  \centering
  \includegraphics[width=.9\linewidth]{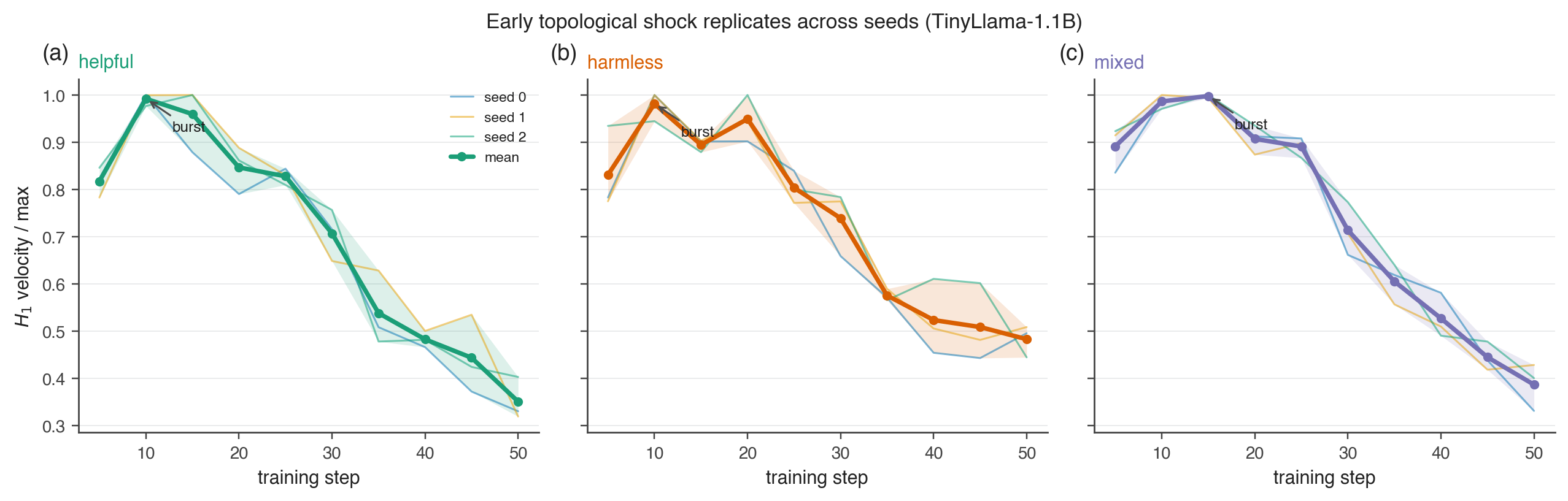}
  \vspace{0.5em}
  \includegraphics[width=.55\linewidth]{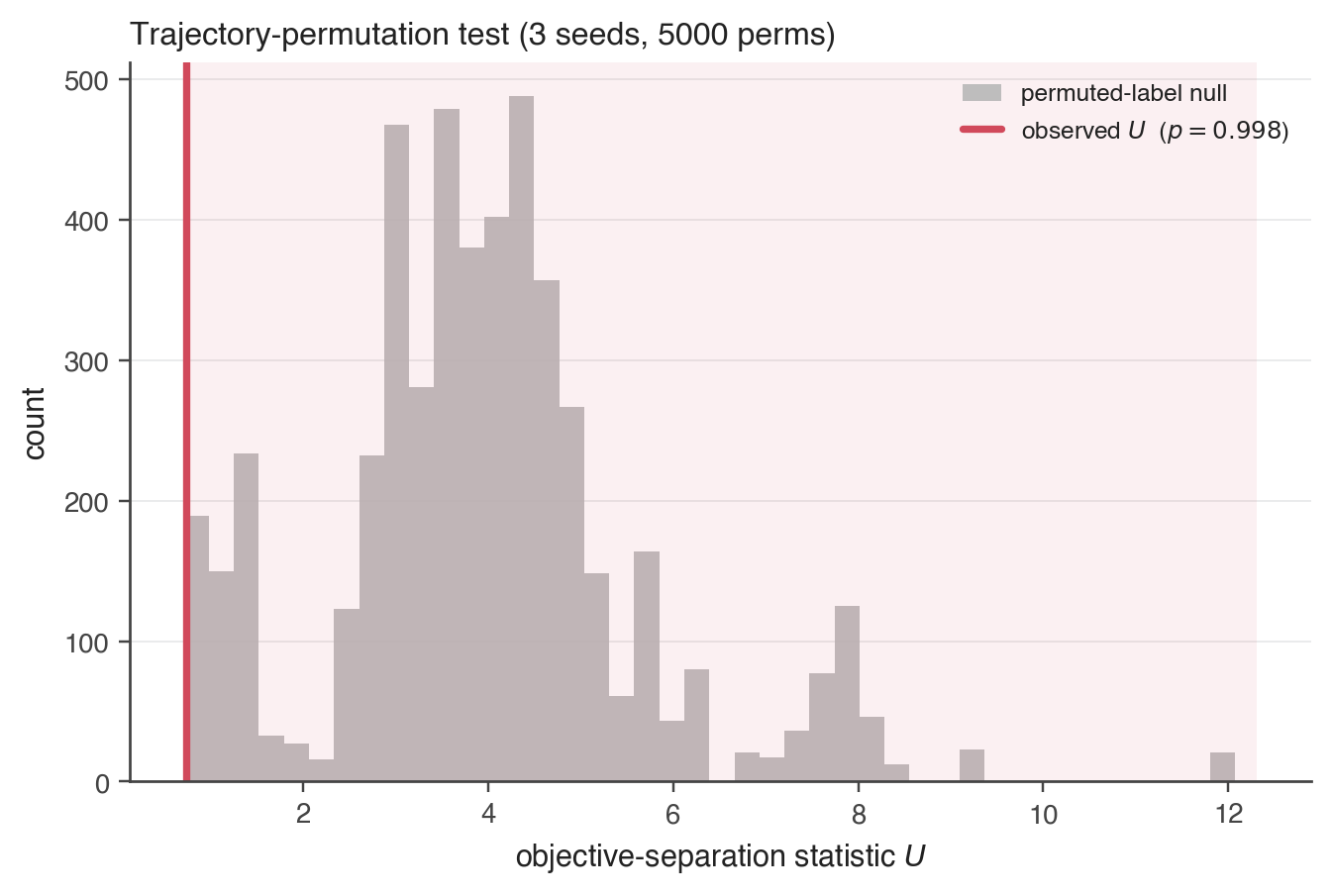}
  \caption{\textbf{Seed replication (TinyLlama-1.1B).} \emph{Top:} early-window $H_1$ velocity overlaid across
  three independent seeds the burst follows nearly the same curve for every objective and seed.
  \emph{Bottom:} the observed objective-separation $U$ falls within the trajectory-permutation null, so
  objectives are not separated during the burst.}
  \label{fig:seedstiny}
\end{figure}

\begin{figure}[p]
  \centering
  \includegraphics[width=.9\linewidth]{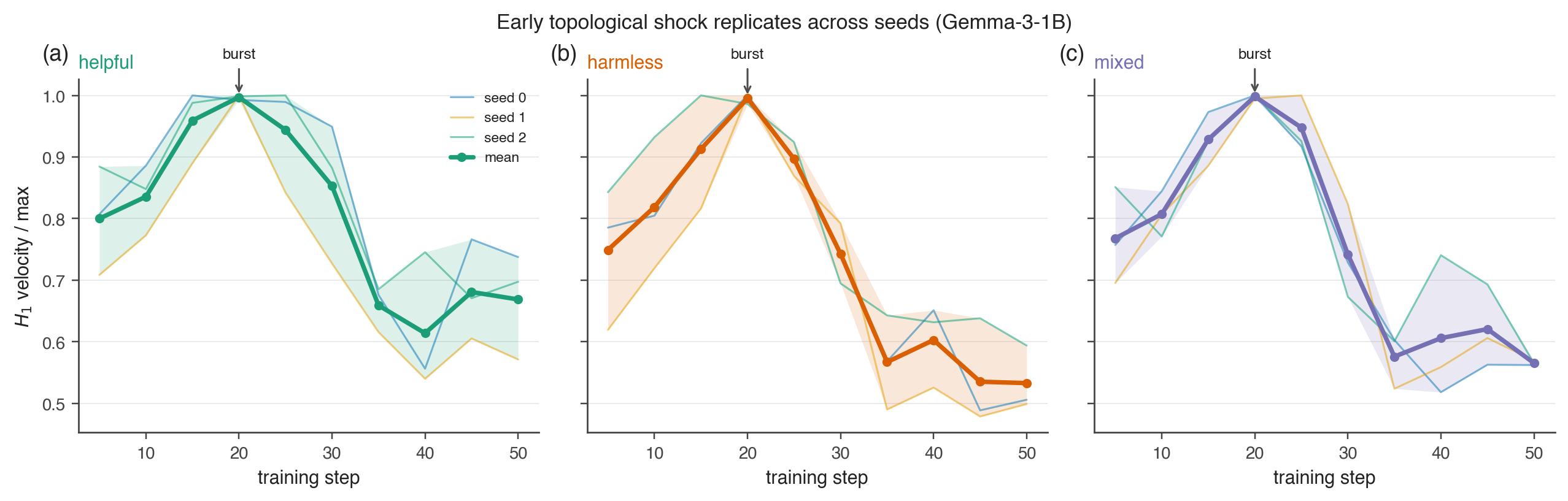}
  \vspace{0.5em}
  \includegraphics[width=0.55\linewidth]{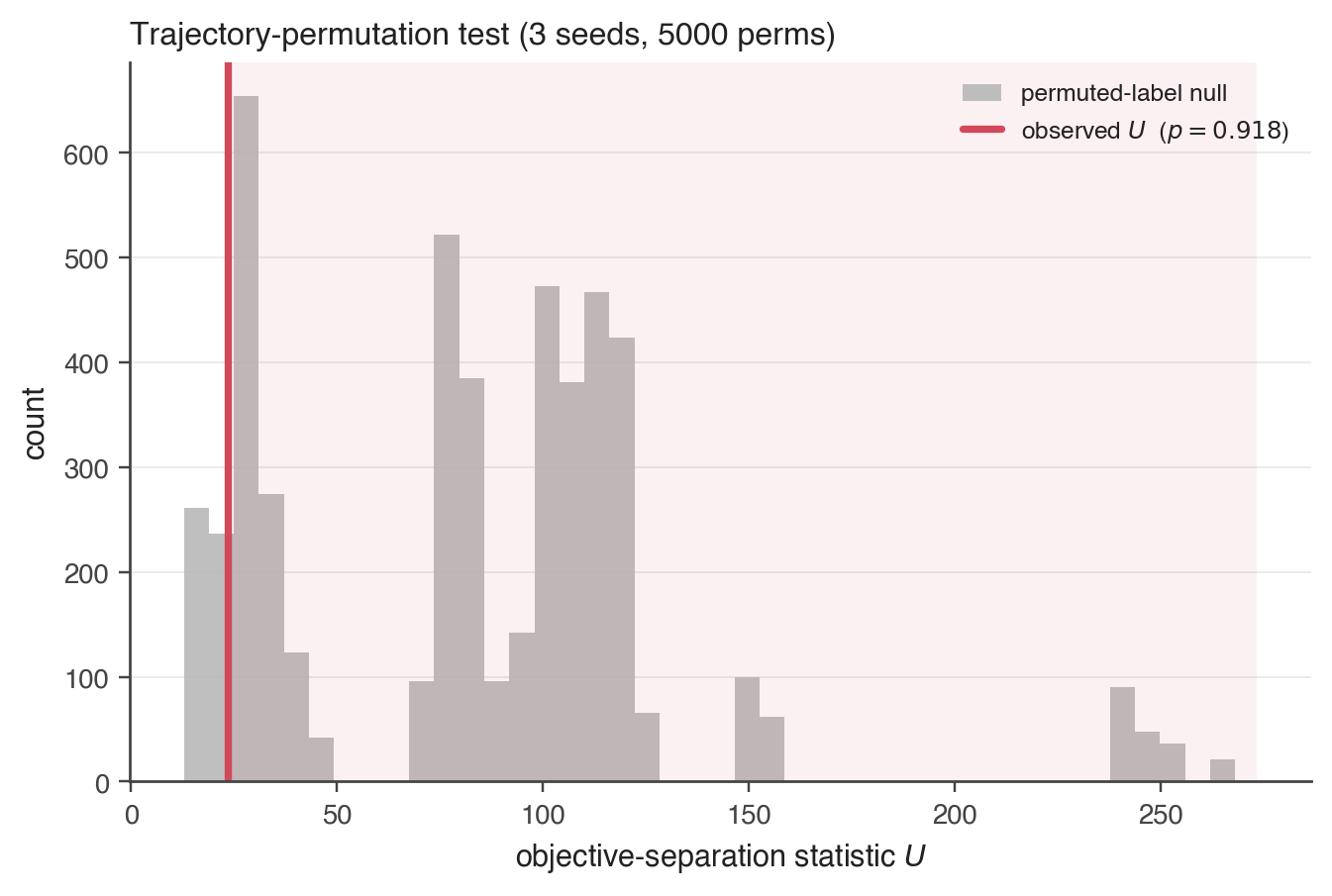}
  \caption{\textbf{Seed replication (Gemma-3-1B).} \emph{Top:} early-window $H_1$ velocity overlaid across
  three independent seeds the burst follows nearly the same curve for every objective and seed.
  \emph{Bottom:} the observed objective-separation $U$ falls within the trajectory-permutation null, so
  objectives are not separated during the burst.}
  \label{fig:seeds}   
\end{figure}

\begin{figure}[p]
  \centering
  \includegraphics[width=.9\linewidth]{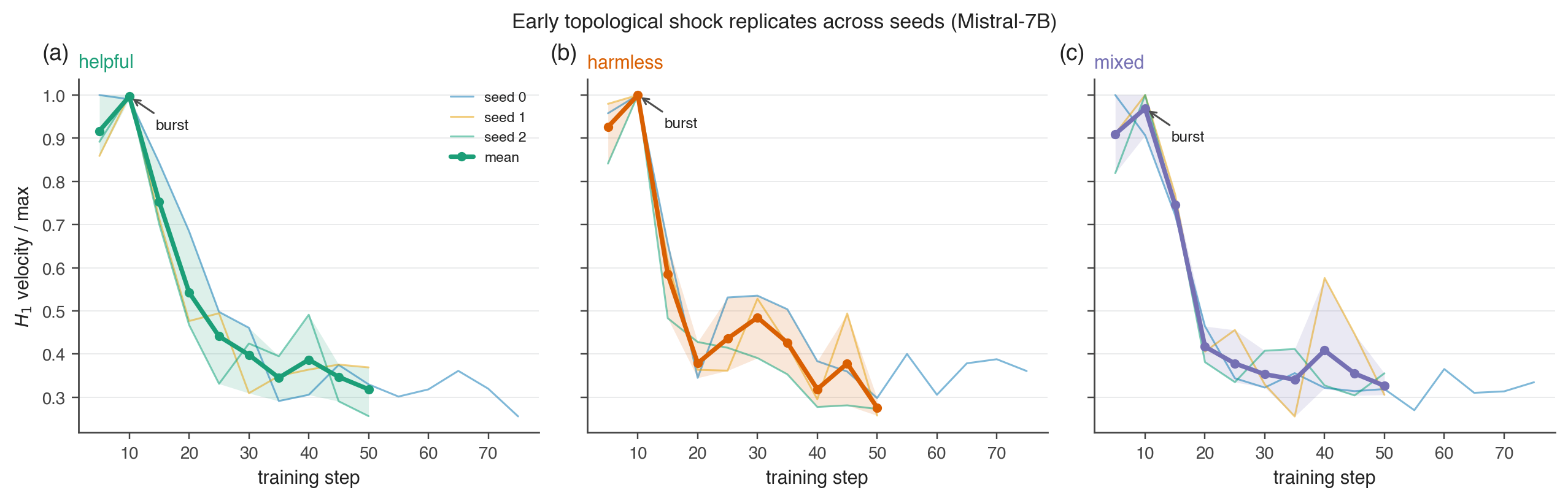}
  \vspace{0.5em}
  \includegraphics[width=0.55\linewidth]{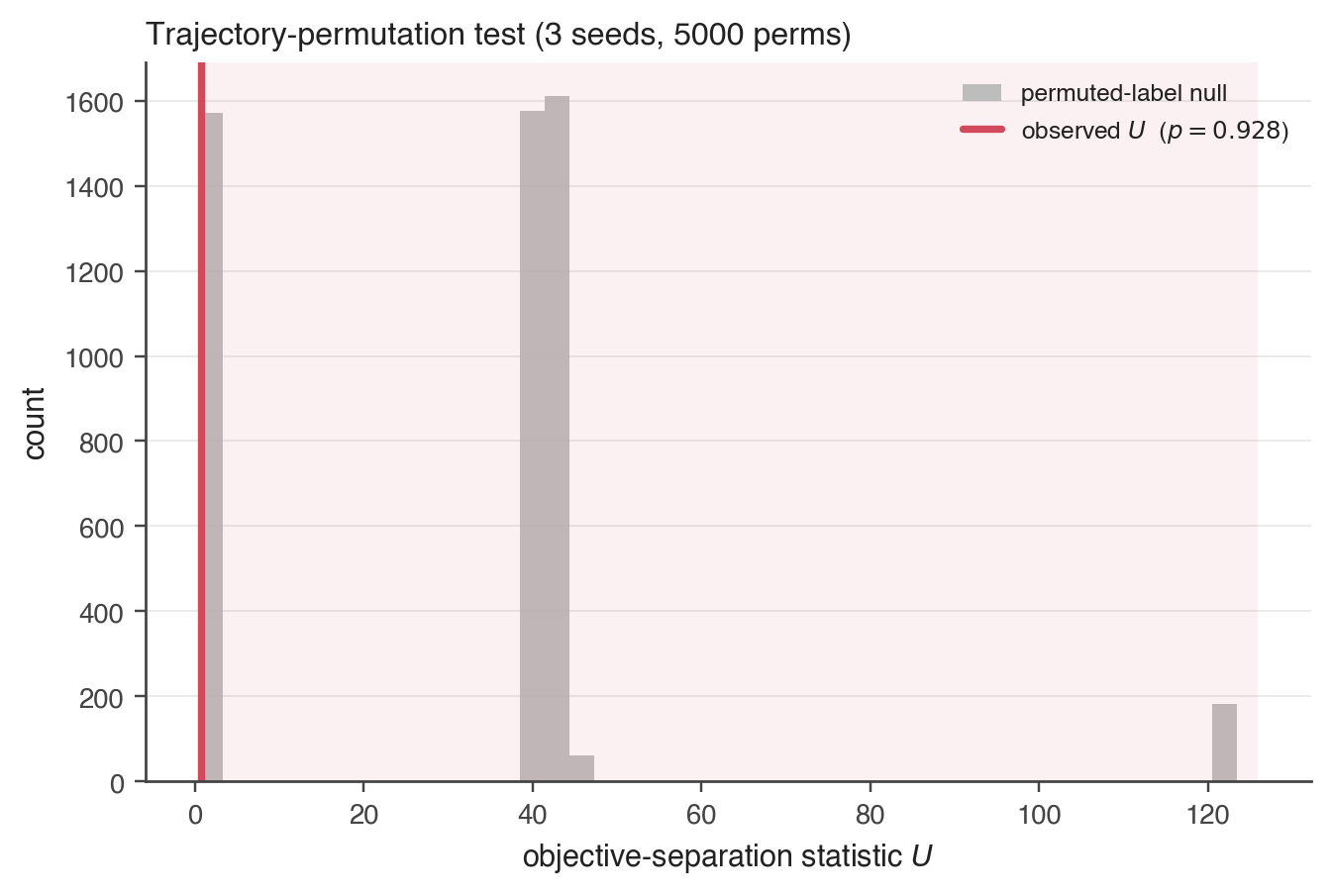}
  \caption{\textbf{Seed replication (Mistral-7B).} \emph{Top:} early-window $H_1$ velocity overlaid across
  three independent seeds the burst follows nearly the same curve for every objective and seed.
  \emph{Bottom:} the observed objective-separation $U$ falls within the trajectory-permutation null, so
  objectives are not separated during the burst.}
  \label{fig:seedsmistral}
\end{figure}

Across TinyLlama-1.1B, Gemma-3-1B, and Mistral-7B, the
early-window $H_1$ velocity follows nearly identical
rise--peak--decay trajectories across seeds and objectives
(Figs.~\ref{fig:seedstiny},
\ref{fig:seeds}, and
\ref{fig:seedsmistral}).

The seed-replicated objective-separation test of
eq.~\eqref{sec:methU} becomes non-degenerate once multiple
trajectories per objective are available.
For all three models, the observed statistic lies comfortably within
its trajectory-permutation null:

\begin{itemize}
\item TinyLlama-1.1B: $U=0.74$, $p=0.998$;
\item Gemma-3-1B: $U=23.6$, $p=0.92$;
\item Mistral-7B: $U=0.76$, $p=0.93$.
\end{itemize}

\begin{figure}[p]
  \centering
  \includegraphics[width=.45\linewidth]{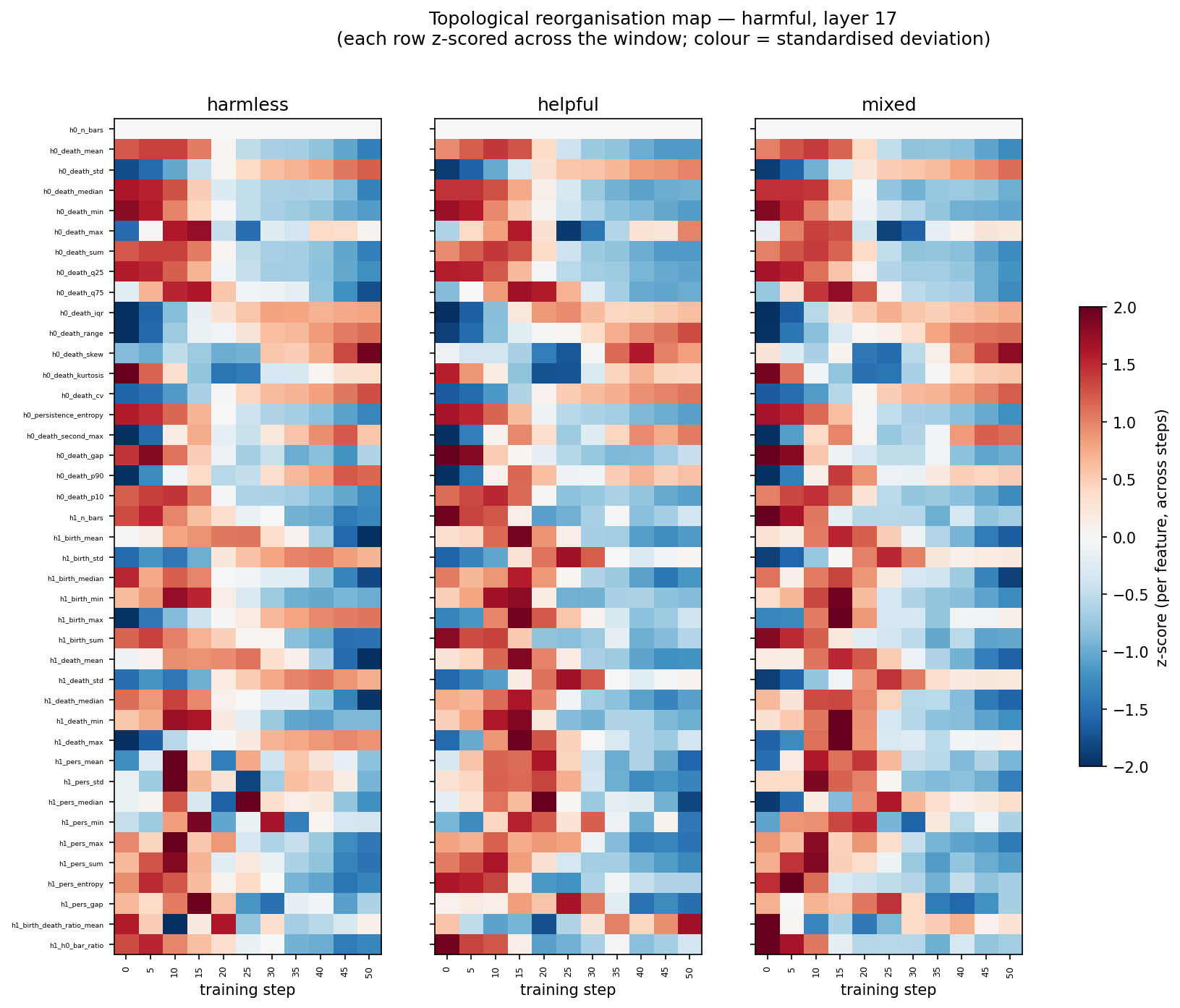}\hfill
  \includegraphics[width=.45\linewidth]{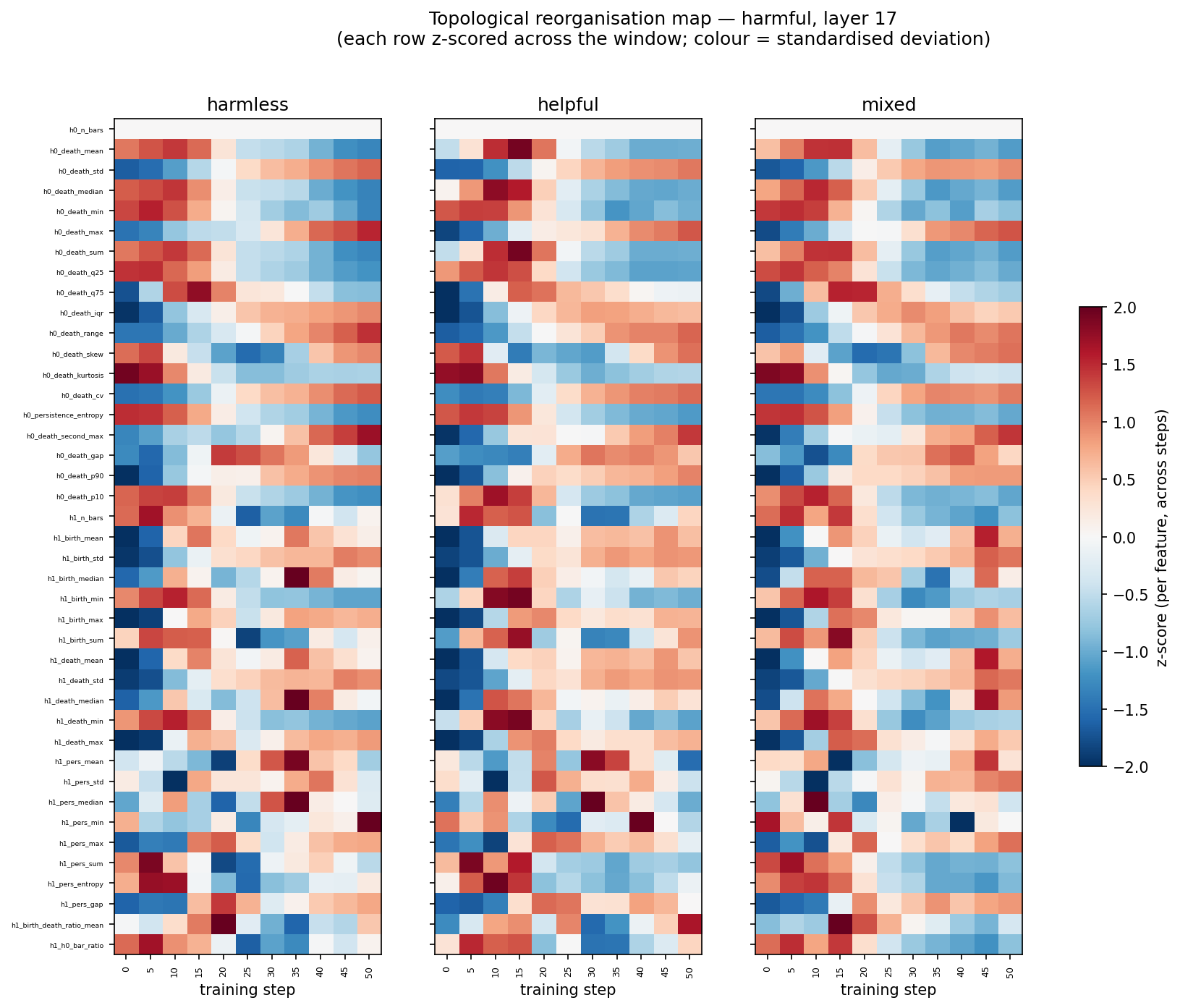}
  \caption{Reorganization maps for the two added TinyLlama seeds (harmless objective), matching the
  early-window structure of the original seed.}
  \label{fig:appseedreorgtiny}
\end{figure}

\begin{figure}[p]
  \centering
  \includegraphics[width=0.45\linewidth]{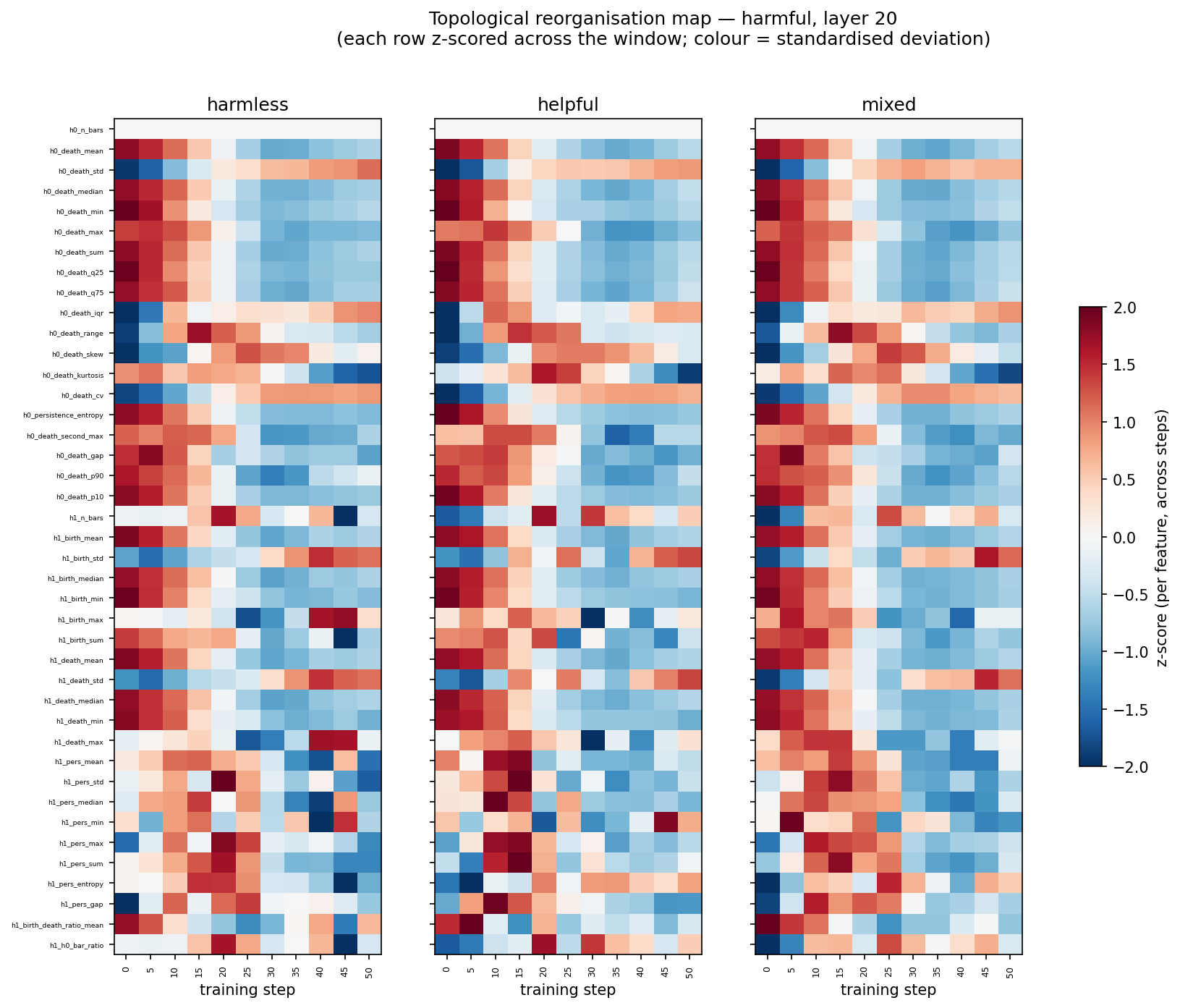}\hfill
  \includegraphics[width=0.45\linewidth]{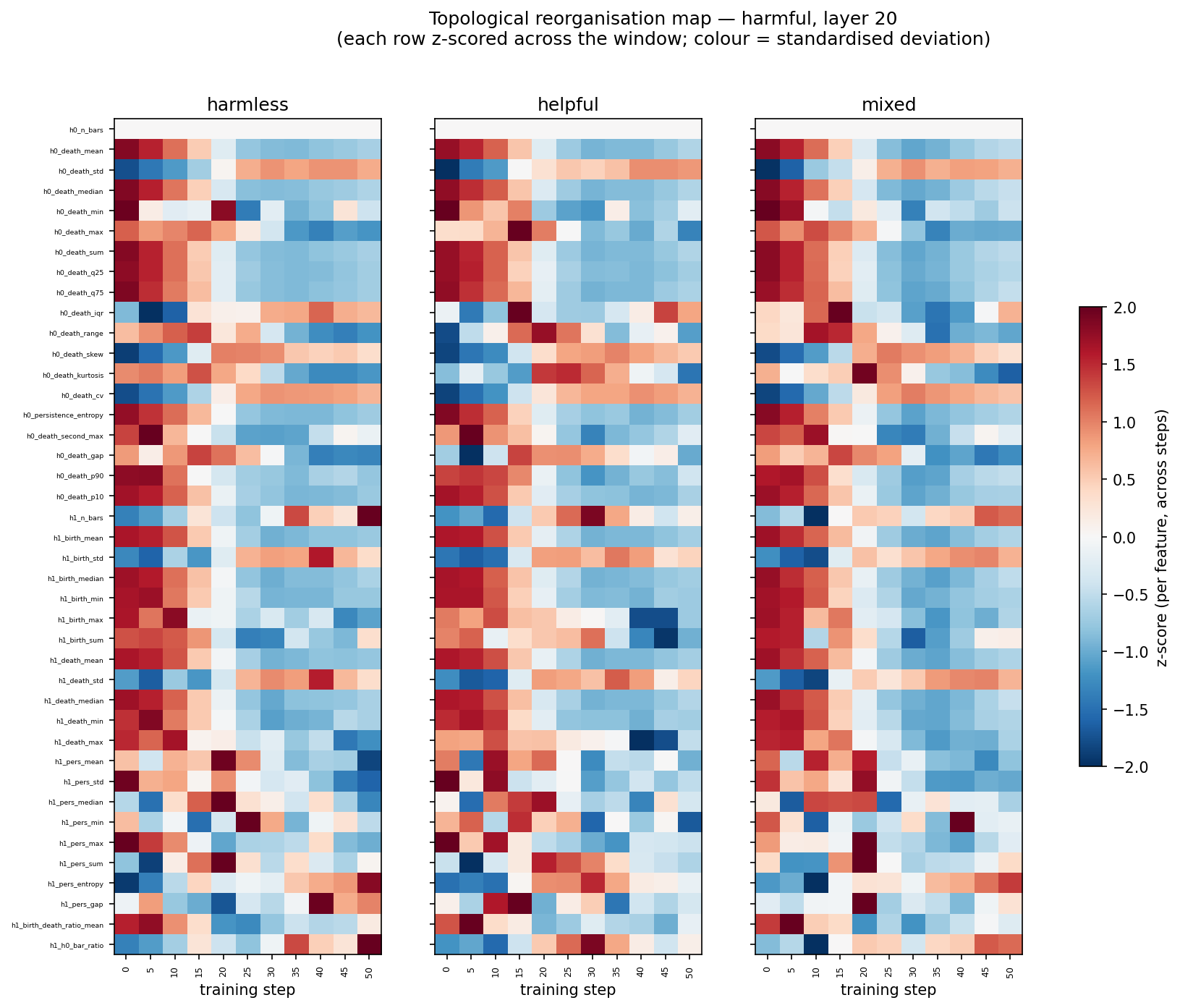}
  \caption{Reorganization maps for the two added Gemma seeds (harmless objective), reproducing the structure
  of the original run.}
  \label{fig:appseedreorg}
\end{figure}

\begin{figure}[p]
  \centering
  \includegraphics[width=0.45\linewidth]{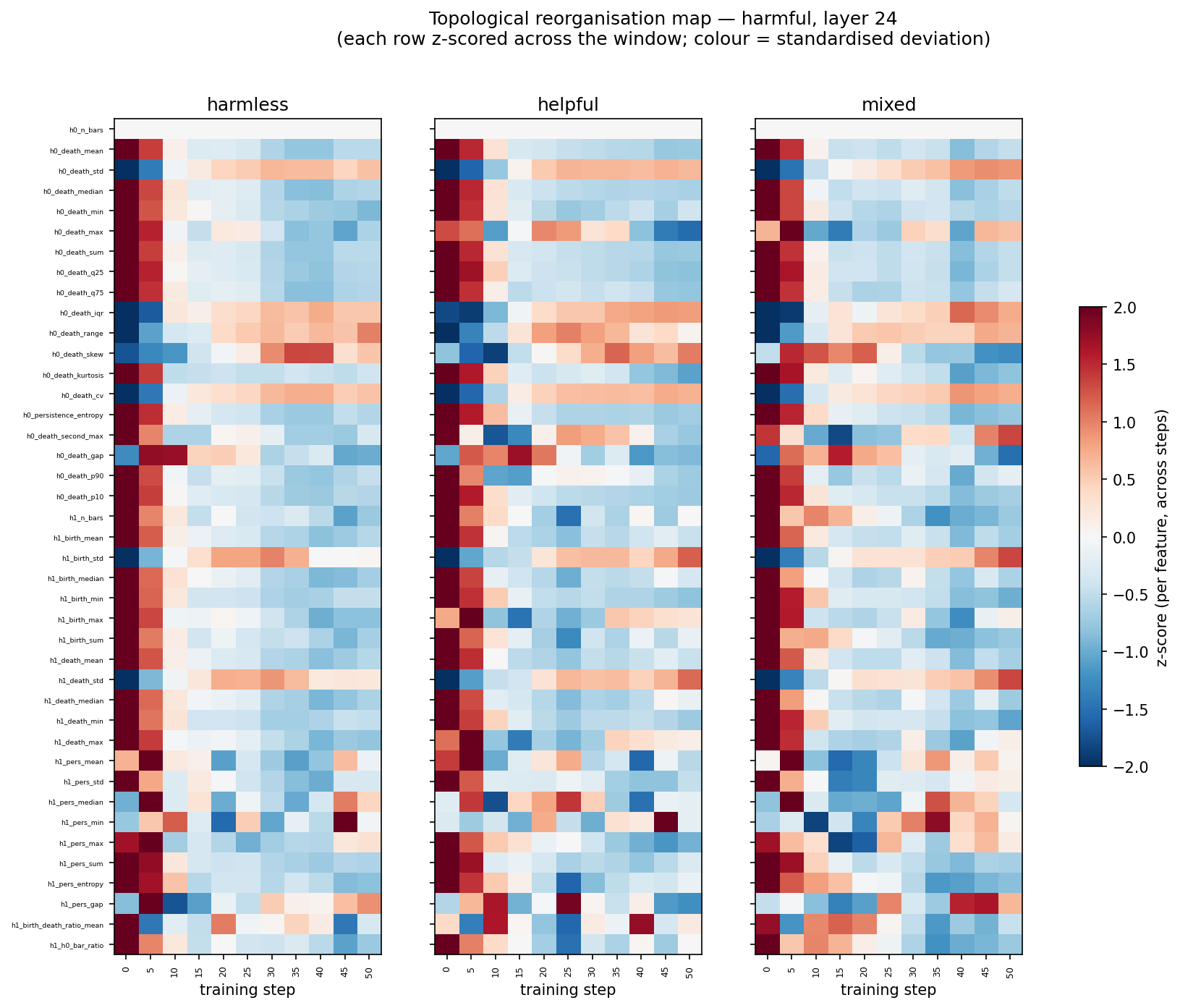}\hfill
  \includegraphics[width=0.45\linewidth]{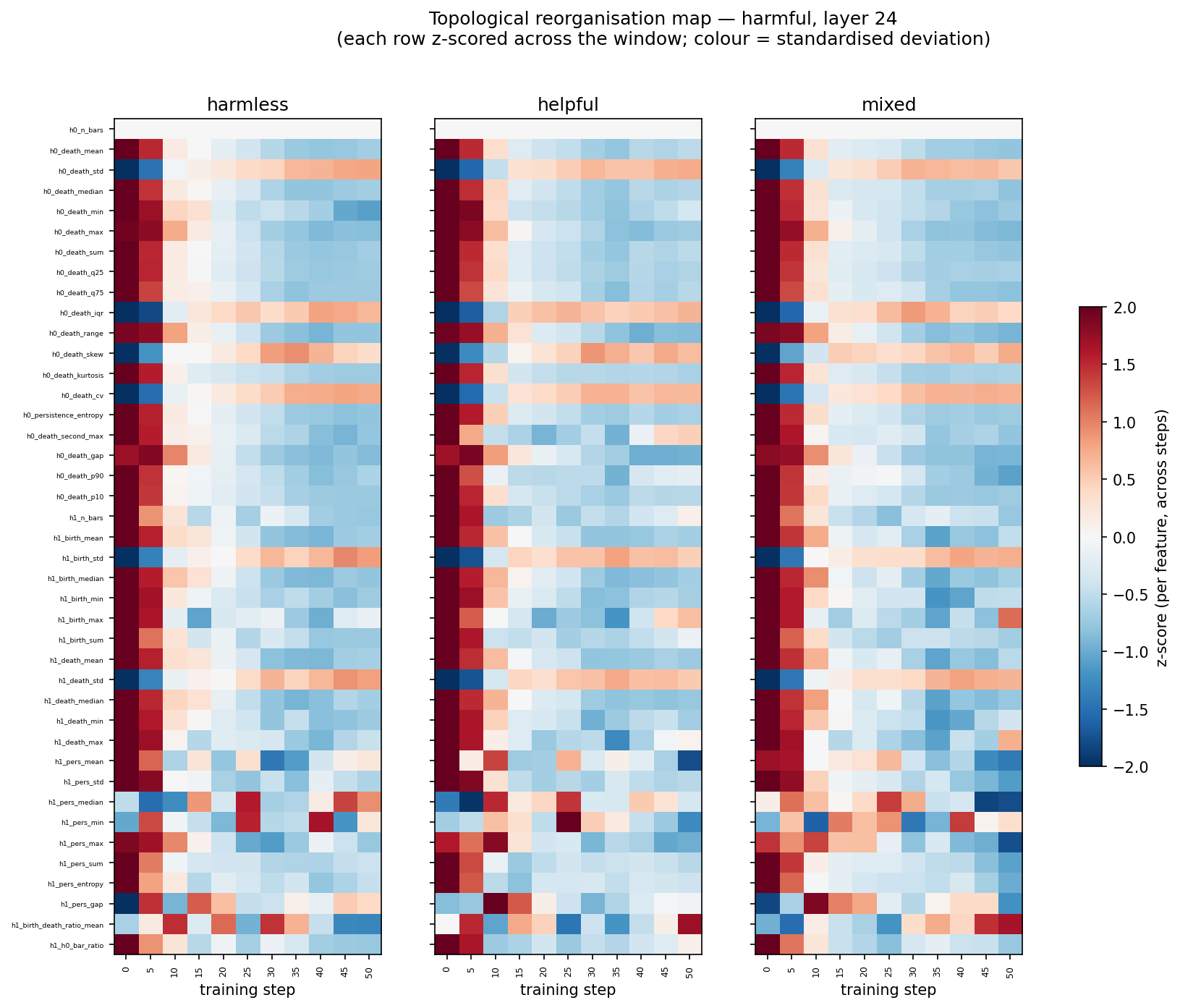}
  \caption{Reorganization maps for the two added Mistral seeds (harmless objective), matching the
  early-window structure of the original run.}
  \label{fig:appseedreorgmistral}
\end{figure}

Thus, the alignment objectives are statistically indistinguishable
during the initial transient burst.
The additional seeds also reproduce the original reorganization maps
(Figs.~\ref{fig:appseedreorgtiny},
\ref{fig:appseedreorg},
and \ref{fig:appseedreorgmistral}),
demonstrating that the early transient phenomenon is robust to both
LoRA initialization and data ordering.

Absolute values of $U$ are not directly comparable across models,
since each model induces its own standardized feature scale.

\subsection{Additional Analyses}
\label{app:misc}

\begin{figure}[b]
  \centering
  \includegraphics[width=0.80\linewidth]{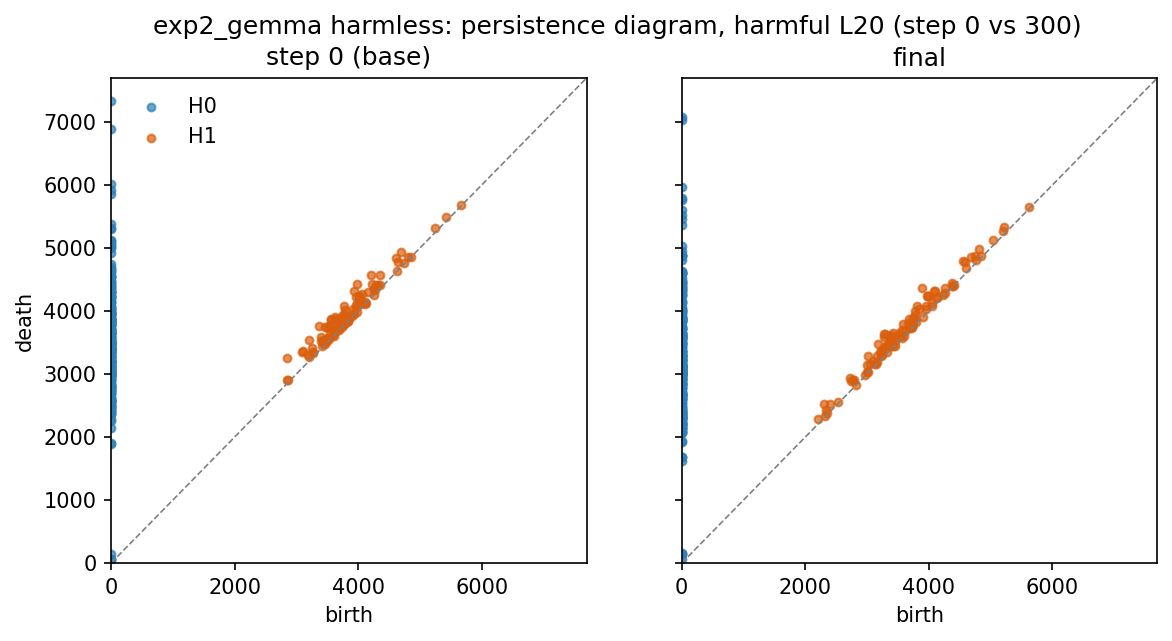}
  \vspace{0.5em}
  \includegraphics[width=0.80\linewidth]{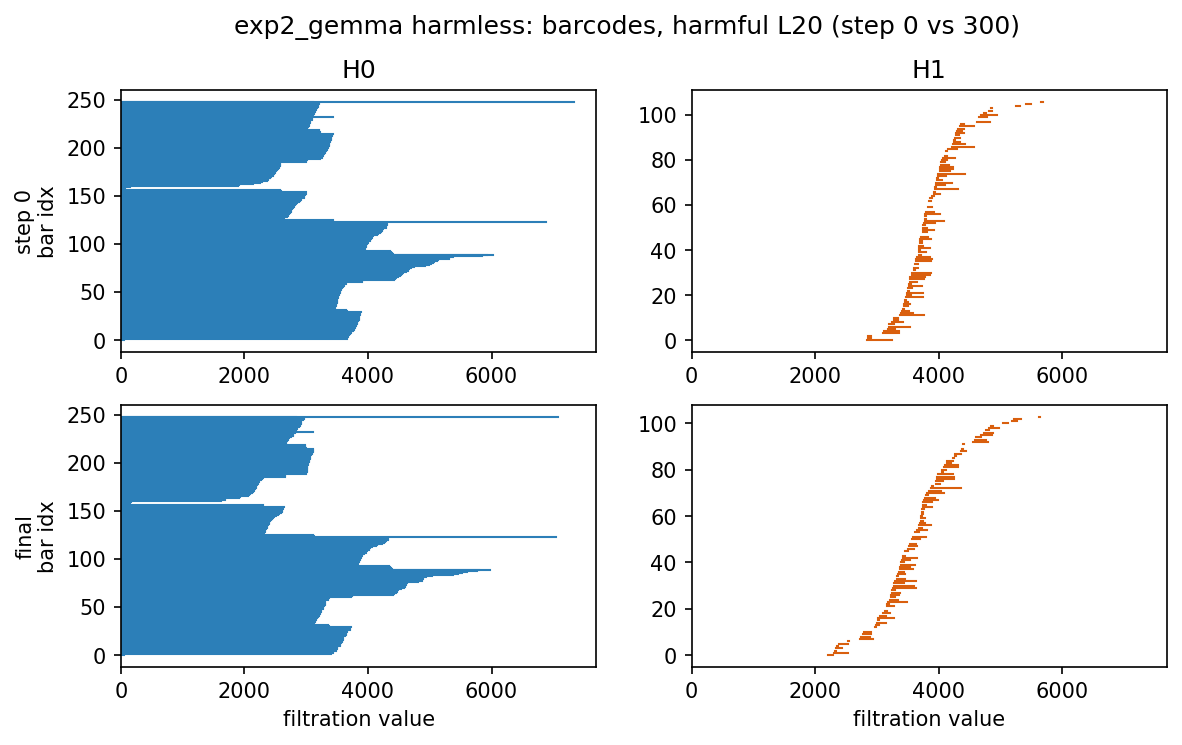}
  \caption{Gemma-3-1B persistence diagram (top) and barcode (bottom), harmless run, layer $20$.
  Fine-tuning mildly shortens and thins the bars.}% The compression the $41$-vector quantifies.}
  \label{fig:appbarcodes}
\end{figure}

\begin{figure}[ptb]
  \centering
  \includegraphics[width=1.00\linewidth]{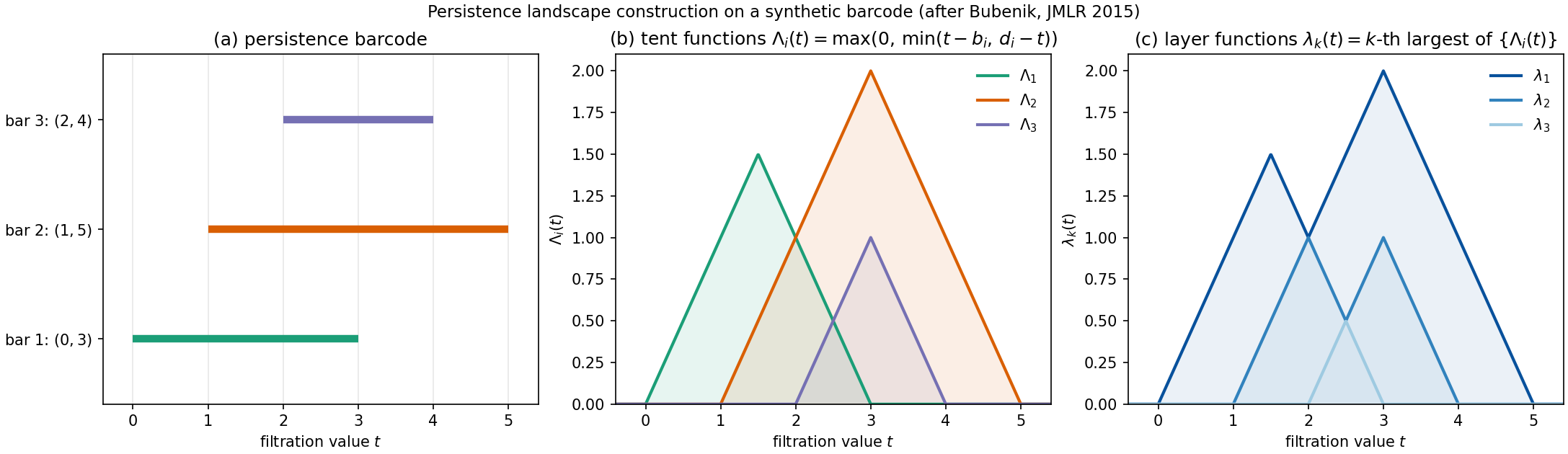}
  \vspace{0.5em}
  \includegraphics[width=1.00\linewidth]{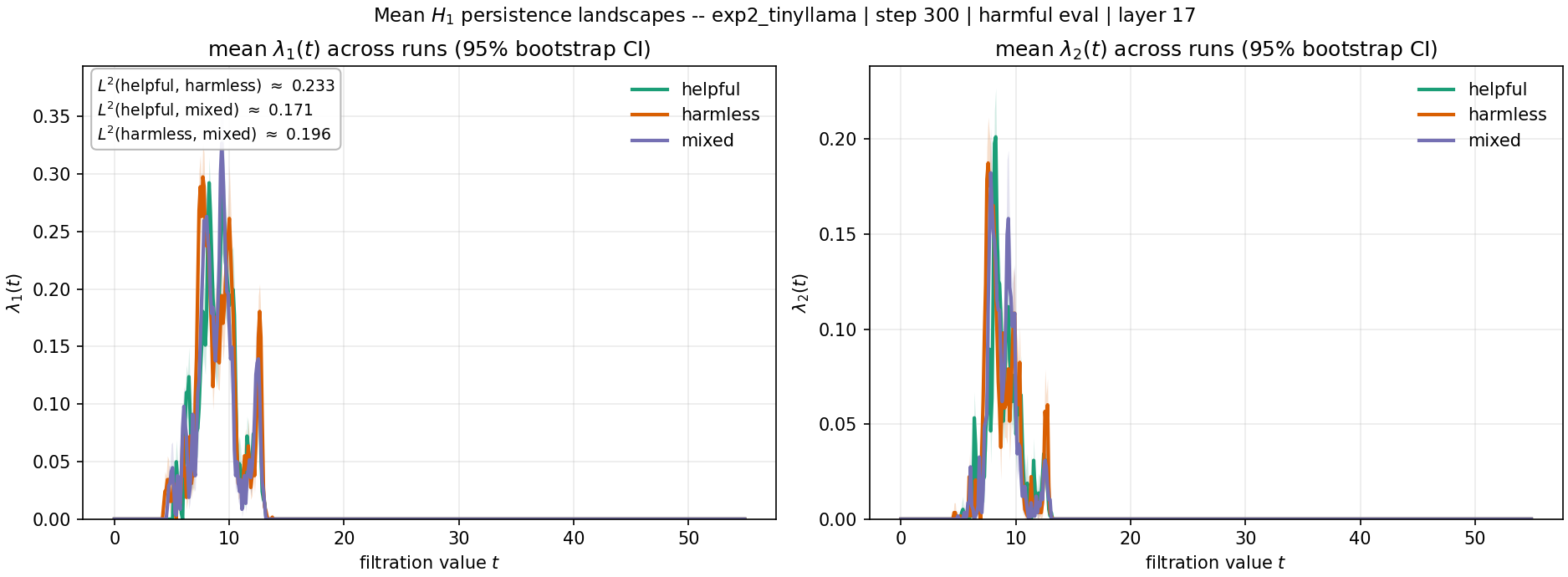}
  \caption{Persistence landscapes: pedagogical construction (top) and the three objectives' mean $H_k$
  landscapes with $95\%$ bootstrap bands and pairwise $L^2$ distances (bottom).}
  \label{fig:applandscape}
\end{figure}

Figure~\ref{fig:appbarcodes} shows representative persistence
diagrams and barcodes for Gemma-3-1B under the harmless objective.
Figure~\ref{fig:applandscape} illustrates persistence landscapes and
their associated bootstrap summaries.

Table~\ref{tab:appshock} reports the checkpoint-order permutation
null for all model-objective combinations.
The observed early concentration exceeds the shuffled null in
$11/12$ cells, with only the helpful run of Gemma-3-1B failing to
reach significance.

Table~\ref{tab:appleadlag} reports the peak-step comparison between
topological and functional velocities.
Topology never lags function and leads by at most one dense-window
checkpoint.

\begin{figure}[ptb]
  \centering
  \includegraphics[width=0.7\linewidth]{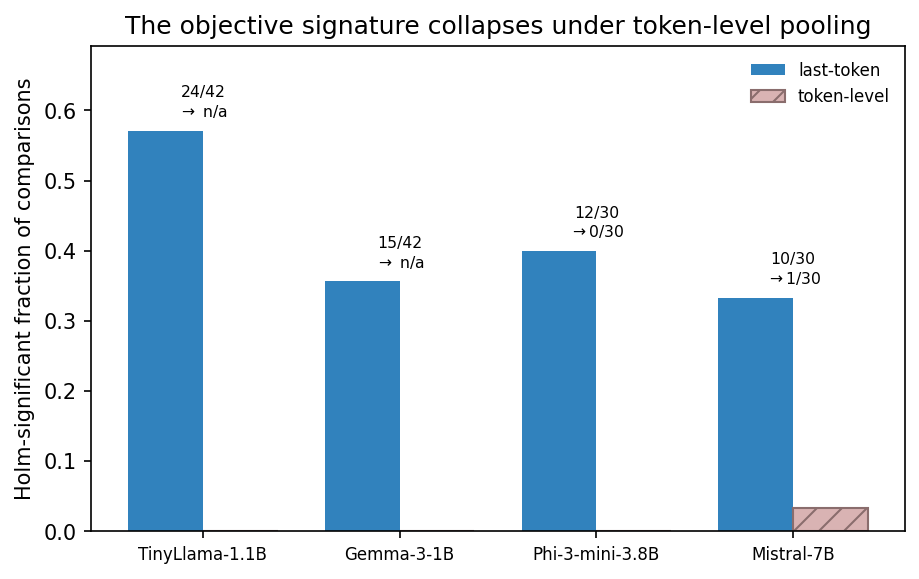}
  \caption{Holm-significant \emph{fraction} of landscape comparisons, last-token vs.\ token-level pooling
  (raw counts annotated). Pooling the last $16$ tokens collapses the objective signal (Phi-3
  $12/30\!\to\!0/30$, Mistral $10/30\!\to\!1/30$), so the discriminating structure sits at the decision
  token. (TinyLlama and Gemma have no token-level variant.)}
  \label{fig:appholm}
\end{figure}

Figure~\ref{fig:appholm} shows that token-level pooling largely
eliminates objective separation, localizing the discriminative
signal to the decision token.

\begin{figure}[ptb]
  \centering
  \includegraphics[width=0.49\linewidth]{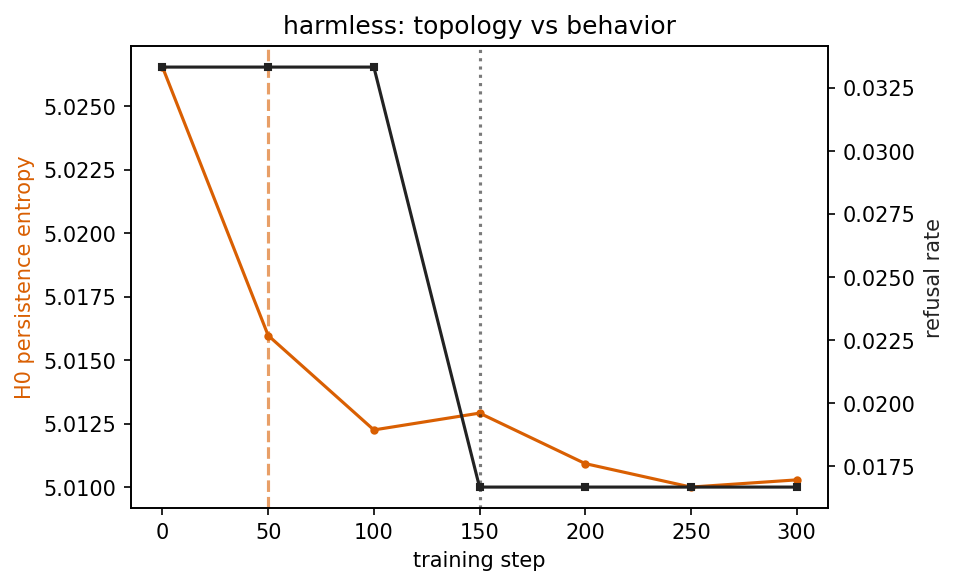}\hfill
  \includegraphics[width=0.49\linewidth]{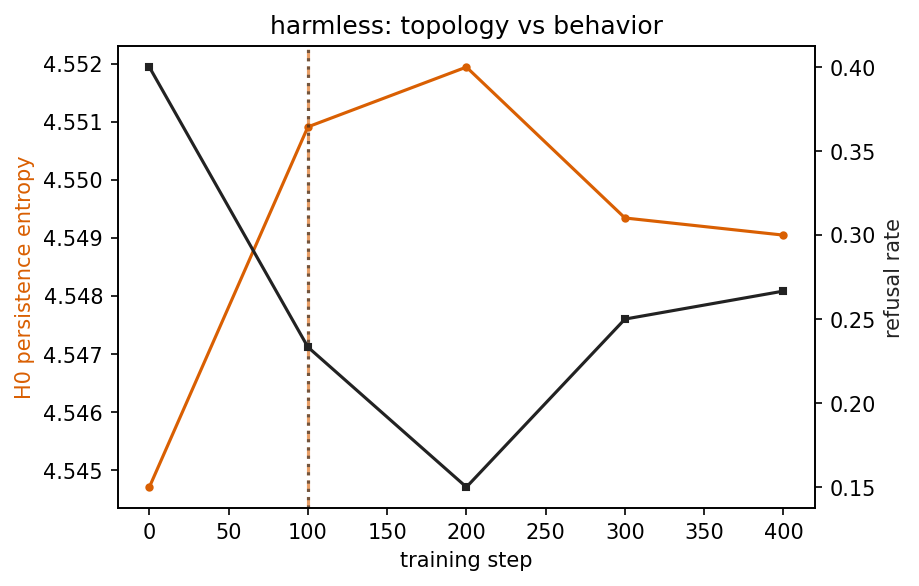}
  \caption{Topology (left axis) vs.\ refusal rate (right axis) over training, harmless run:  TinyLlama (base, left) and Phi-3 (instruct, right). Refusal is pinned at the noise floor for the base model and \emph{moves} only for the instruction-tuned Phi-3.}
  \label{fig:appgrok}
\end{figure}

Figure~\ref{fig:appgrok} gives topology-versus-refusal overlays for
representative base and instruction-tuned models.

\begin{figure}[ptb]
  \centering
  \includegraphics[width=\linewidth]{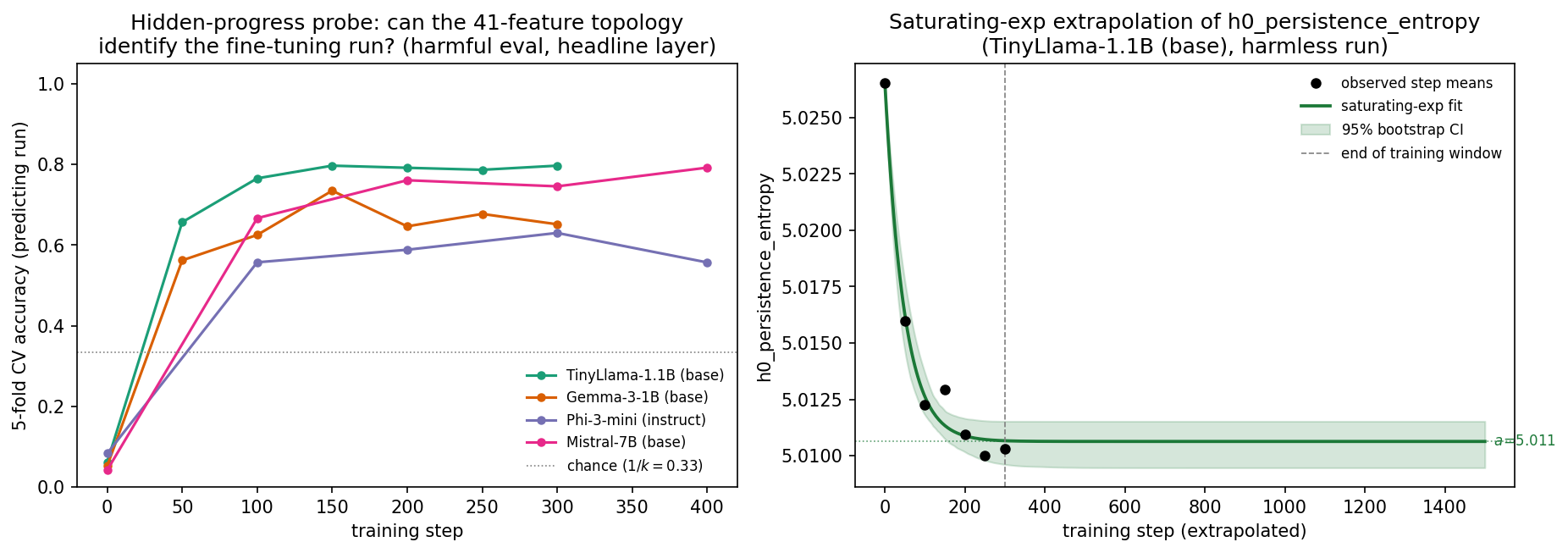}
  \caption{\emph{Left:} a $5$-fold CV probe on the $41$-vector becomes run-discriminative by step $\sim\!50$--$100$ even where refusal is flat (hidden progress). \emph{Right:} saturating-exponential fit to $h_0$ persistence entropy.}
  \label{fig:appfollow}
\end{figure}

\begin{figure}[ptb]
  \centering
  \includegraphics[width=0.85\linewidth]{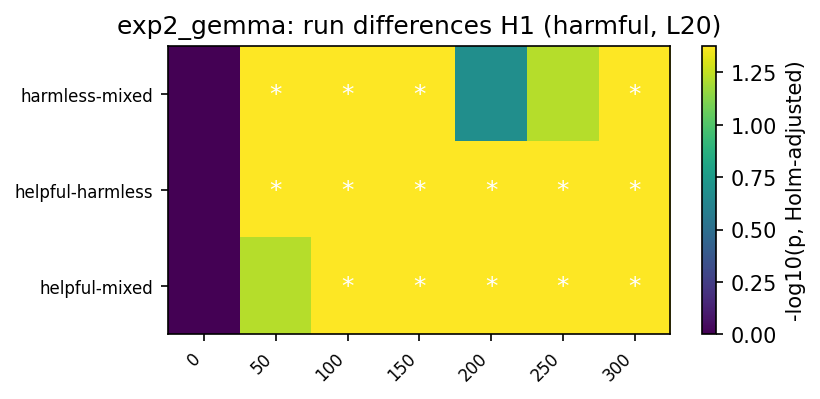}\hfill
  \caption{The subsample-level Holm-adjusted landscape-permutation heatmap (Gemma, $H_1$). The inflated test the cell-level analysis replaces.}
  \label{fig:apppca}
\end{figure}

Finally,
Fig.~\ref{fig:appfollow} presents the hidden-progress probe and
saturating-exponential fit to $H_0$ persistence entropy, while
Fig.~\ref{fig:apppca} shows the exploratory subsample-level
landscape-permutation heatmap discussed in
\S\ref{sec:res-floor}.

\begin{figure}[p]
  \centering
  \includegraphics[width=\linewidth]{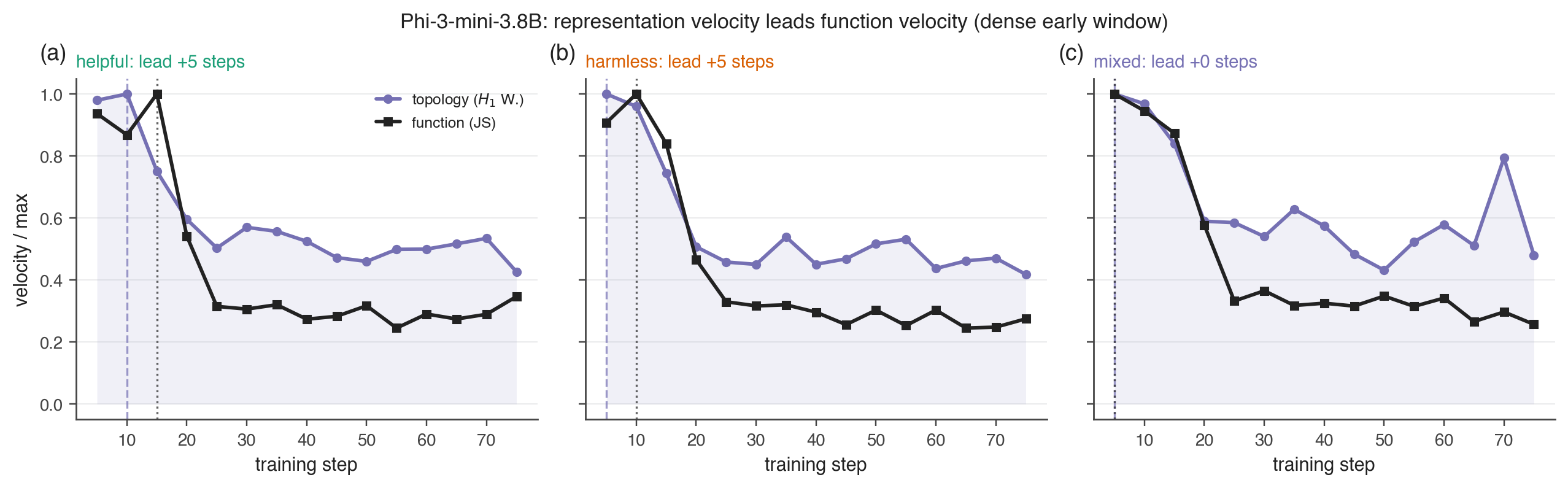}
  \caption{\textbf{Representation leads function (Phi-3, dense).} $H_1$ Wasserstein velocity (topology,
  purple) and JS velocity of the output distribution (function, black) on a shared checkpoint axis. Both
  peak in the first $5$-$15$ steps with topology marginally first (dashed vs.\ dotted markers). Topology then retains a longer tail. Phi-3 is the model whose refusal genuinely moves, so this is the sharpest available timing test.}
  \label{fig:leadlag}
\end{figure}

\begin{figure}[p]
  \centering
  \includegraphics[width=\linewidth]{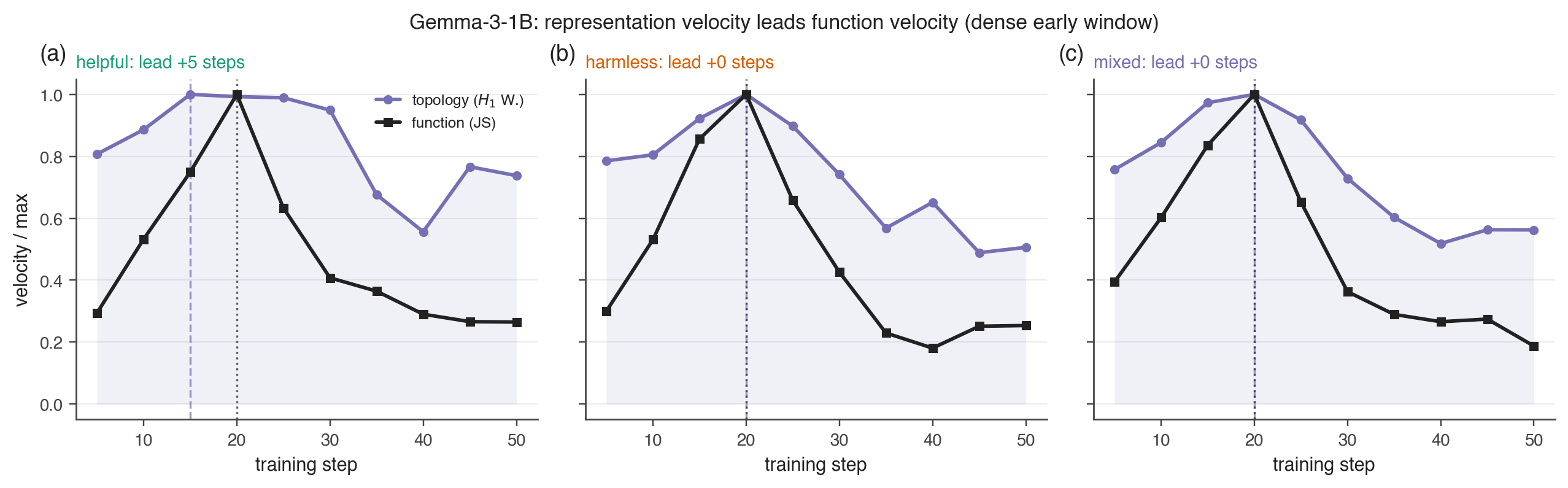}
  \caption{Gemma-3-1B representation ($H_1$ Wasserstein) vs.\ function (JS) velocity companion to
  Fig.~\ref{fig:leadlag}. Both peak near step $20$. Topology leads by at most one checkpoint.}
  \label{fig:appleadgemma}
\end{figure}

\section{Compute and Implementation Details}
\label{app:compute}

\subsection{Computational Environment}

\paragraph{Hardware.}
Experiments were conducted on a single Apple M3 Max workstation with a $16$-core CPU, integrated $40$-core GPU, and $64$ GB unified memory running macOS~26. Model training and extraction use the Metal Performance Shaders (MPS) backend through PyTorch with \texttt{bfloat16} precision and scaled-dot-product attention (\texttt{sdpa}). Device selection follows the hierarchy \texttt{mps}$>\texttt{cuda}>\texttt{cpu}$, allowing the same codebase to execute unchanged on CUDA-enabled systems.

\paragraph{Software.}
The training and extraction pipeline uses PyTorch, Transformers, PEFT, Accelerate, and Datasets. Persistent homology computations are performed using \texttt{ripser}, while persistence distances are computed with \texttt{persim}. Statistical analyses and visualizations use NumPy, SciPy, pandas, scikit-learn, and Matplotlib. The topology and statistics modules are deliberately implemented independently of deep-learning frameworks.

\paragraph{Reproducibility.}
The software environment is managed using \texttt{uv} (Python~3.11). Every run records the random seed, model revision, and software versions alongside the extracted activations. Persistence diagrams are cached per (run, condition, layer) tuple, allowing all analyses to be reproduced directly from stored activations and diagrams without repeating model training or extraction. Code, configurations, and anonymized software accompany the submission.

\subsection{Models and Fine-Tuning}

We study four open-weight decoder-only language models spanning approximately $1$B--$7$B parameters:

\begin{itemize}
\item TinyLlama-1.1B (\texttt{TinyLlama/TinyLlama-1.1B-intermediate-step-1431k-3T}),
\item Gemma-3-1B (\texttt{google/gemma-3-1b-pt}),
\item Phi-3-mini-3.8B (\texttt{microsoft/Phi-3-mini-4k-instruct}), and
\item Mistral-7B (\texttt{mistralai/Mistral-7B-v0.1}).
\end{itemize}

Phi-3-mini-3.8B is instruction-tuned, while the remaining models are pretrained base models.

Each model is fine-tuned by supervised fine-tuning on the preferred (\emph{chosen}) responses of the helpful, harmless, and mixed HH-RLHF subsets \citep{bai2022training}. We use LoRA with rank $r=8$, $\alpha=16$, dropout $0.05$, and \texttt{task\_type=CAUSAL\_LM}, applied to the attention projections. Optimization uses a learning rate of $2\times10^{-4}$, micro-batches of size $4$, gradient accumulation $8$ (effective batch size $32$), and a maximum sequence length of $256$. All objectives use identical hyperparameters and random seeds.

The $1$B models are trained for $300$ steps on $3000$ examples with checkpoints every $50$ steps. The larger models are trained for $400$ steps on $6000$ examples with checkpoints every $100$ steps. To resolve the earliest stages of alignment, we additionally perform a dense early-window analysis that checkpoints every $5$ steps up to step $50$ while keeping the same optimization schedule.

\subsection{Activation Extraction and Persistent Homology}

At each checkpoint, we evaluate the model on a fixed evaluation set consisting of $250$ prompts per condition and extract the final-token hidden state at five evenly spaced transformer layers. This yields one activation point cloud of $n=250$ points for each (condition, layer) pair.

For each cloud, we draw $B=64$ overlapping subsamples of size $m=160$ and compute Vietoris--Rips persistent homology up to dimension one using the Euclidean metric. Persistent homology is computed with \texttt{ripser}, and topological velocities are measured using order-$2$ Wasserstein distances between consecutive-checkpoint persistence diagrams via \texttt{persim}. Unless otherwise stated, analyses are performed on the resulting barcode summaries rather than on the raw diagrams.

The behavioral analyses use the same saved checkpoints and evaluation prompts. Refusal rates and response lengths are computed on a fixed $60$-prompt subset with at most $64$ generated tokens.

\subsection{The 41-Dimensional Barcode Summary}

Each persistence diagram is reduced to a fixed-length vector $\mathbf z\in\mathbb R^{41}$ using the descriptor family of \citet{fay2026shape}. All statistics are computed on finite bars only, with the essential infinite $H_0$ bar removed.

The summary consists of $19$ statistics describing $H_0$, $21$ statistics describing $H_1$, and one cross-degree feature:

\begin{itemize}
\item \textbf{$H_0$ ($19$):} number of bars, summary statistics of component death times, persistence entropy, and salience measures based on the largest and second-largest deaths;
\item \textbf{$H_1$ ($21$):} number of bars together with summary statistics of loop birth times, death times, and persistences, persistence entropy, salience measures, and the mean birth/death ratio;
\item \textbf{Cross ($1$):} the ratio $n_{H_1}/n_{H_0}$.
\end{itemize}

Persistence entropy is defined by
$$
-\sum_i p_i\log p_i,
\qquad
p_i=\frac{\ell_i}{\sum_j\ell_j},
$$
where $\ell_i$ denotes the length of the $i$th bar. Together, these descriptors summarize changes in component mergers, loop structure, and persistence distributions in a form that is directly comparable across checkpoints and objectives.

\section{Additional Statistical Details and Robustness Checks}
\label{app:stats}

\subsection{Statistical Inference and Robustness}

\paragraph{Cell-level inference.}
Each activation cloud is represented by $B=64$ overlapping subsamples. These subsamples provide an estimate of variability arising from finite prompt sets but do not constitute independent observations. Consequently, all confirmatory statistical inference is performed at the \emph{cell level}, where a cell corresponds to a fixed combination of model, objective, evaluation condition, layer, and checkpoint. Subsample-level analyses are used only for exploratory visualizations, such as PCA projections and persistence-landscape maps.

\paragraph{Effect sizes and confidence intervals.}
Our primary measures of difference are per-feature effect sizes (Hedges' $g$ and Cliff's $\delta$) and the multivariate energy distance. Ridge-regularized Mahalanobis distances were computed as a cross-check and yielded qualitatively similar conclusions. All effect sizes are reported with percentile bootstrap $95\%$ confidence intervals obtained from at least $2000$ bootstrap resamples of the appropriate experimental unit.

\paragraph{Trajectory-level permutation testing.}
The objective-separation statistic $U$ is assessed by permuting whole trajectories rather than individual subsamples. With only one trajectory per objective, permutation testing is degenerate because the all-pairs trajectory dispersion is invariant under relabeling. We therefore perform three-seed replications for TinyLlama-1.1B, Gemma-3-1B, and Mistral-7B, producing nine trajectories per model.

Under the null hypothesis that objective labels are exchangeable, we consider all size-preserving assignments of these nine trajectories into three groups of size three. The number of such assignments is
\[
\frac{9!}{(3!)^3}=1680,
\]
yielding an exact permutation test with minimum attainable $p$-value
\[
p_{\min}=\frac{1}{1681}\approx5.95\times10^{-4},
\]
using the Phipson--Smyth correction
\[
p=
\frac{
1+\#\{T_{\mathrm{perm}}\ge T_{\mathrm{obs}}\}
}{
1+N_{\mathrm{perm}}
}.
\]

\paragraph{Checkpoint-order null and behavioral velocity.} \label{app:js_velocity}
To assess whether topological reorganization is genuinely concentrated early in training, we randomly permute checkpoint order and recompute both the topological velocity and early-concentration statistic. Behavioral velocity is defined analogously using the Jensen--Shannon divergence between consecutive next-token distributions:
\[
\beta_t
=
\overline{
\mathrm{JS}\!\left(
P_t(\cdot|x),
P_{t-1}(\cdot|x)
\right)
}
\]
where the overline denotes averaging over evaluation prompts. Jensen--Shannon divergence is used because it is symmetric, bounded, and remains finite even when consecutive checkpoints assign probability mass to disjoint sets of tokens.

\paragraph{Additional robustness analyses.}
We additionally compare persistent-homology summaries with conventional geometric statistics computed on the same activation clouds, namely centroid drift, total variance, and mean pairwise distance. We further compare the observed signatures with isotropic Gaussian point clouds matched in sample size and ambient dimension and reconstruct activation clouds using the final sixteen token states of each prompt to assess sensitivity to token selection. These analyses are reported in Appendix~\ref{app:controls}.

\paragraph{Multiple comparisons.}
For exploratory subsample-level landscape comparisons, we control the family-wise error rate using the Holm--Bonferroni procedure. Because the corresponding tests treat heavily overlapping subsamples as independent observations, these corrected $p$-values are regarded as exploratory only. Throughout the paper, we therefore use effect sizes for confirmatory inference and reserve $p$-values primarily for null and control tests.

\end{document}

%% file: figures/vietoris-rips-sketch.tex
\tikzstyle{vertex}=[circle, draw=black!85, fill=black!85, thin, scale=0.5]
\tikzstyle{edge}=[line width=1pt]
\tikzstyle{twosimplex}=[edge,draw,fill=blue!50]
\tikzstyle{bar}=[line width=5pt]

\begin{tikzpicture}
\begin{scope}[scale=.7]
\node[vertex] (01) at (0,1) {};
\node[vertex] (02) at (-2,0) {};
\node[vertex] (03) at (3,0.5) {};
\node[vertex] (04) at (-1.8,-2) {};
\node[vertex] (05) at (.2,-1.5) {};

\draw (01) circle (11mm);
\draw (02) circle (11mm);
\draw (03) circle (11mm);
\draw (04) circle (11mm);
\draw (05) circle (11mm);

\draw[edge] (01) -- (02);
\draw[edge] (02) -- (04);
\draw[edge] (04) -- (05);
\end{scope}

\begin{scope}[xshift=9cm,scale=.7]
\node[vertex] (11) at (0,1) {};
\node[vertex] (12) at (-2,0) {};
\node[vertex] (13) at (3,0.5) {};
\node[vertex] (14) at (-1.8,-2) {};
\node[vertex] (15) at (.2,-1.5) {};

\draw (11) circle (13mm);
\draw (12) circle (13mm);
\draw (13) circle (13mm);
\draw (14) circle (13mm);
\draw (15) circle (13mm);

\draw[edge] (11) -- (12);
\draw[edge] (11) -- (15);
\draw[edge] (12) -- (14);
\draw[edge] (14) -- (15);
\end{scope}

\begin{scope}[xshift=18cm,scale=.7]
\node[vertex] (21) at (0,1) {};
\node[vertex] (22) at (-2,0) {};
\node[vertex] (23) at (3,0.5) {};
\node[vertex] (24) at (-1.8,-2) {};
\node[vertex] (25) at (.2,-1.5) {};

\draw[twosimplex] (21.center) -- (22.center) -- (24.center) -- (25.center) -- (21.center) -- cycle;

\draw (21) circle (16mm);
\draw (22) circle (16mm);
\draw (23) circle (16mm);
\draw (24) circle (16mm);
\draw (25) circle (16mm);

\draw[edge] (21) -- (23);
\draw[edge] (22) -- (25);
\end{scope}

\begin{scope}[yshift=-4.5cm]

\node at (0,0) {\huge\bfseries +};
\node at (9cm,0) {\huge\bfseries +};
\node at (18cm,0) {\huge\bfseries +};
\draw[line width=2pt,->] (-5,0) -- (24cm,0) node[anchor=west,xshift=1em] {\Huge $\varepsilon$};

\draw[bar,blue  ]         (-2 ,1  ) -- (21cm,1  );
\draw[bar,blue,dashed]    (-4 ,1  ) -- (23cm,1  );
\draw[bar,blue  ]         (-2 ,0.5) -- (18cm,0.5);
\draw[bar,blue,dashed]    (-4 ,0.5) -- (-2  ,0.5);
\draw[bar,orange]         (9cm,1.5) -- (18cm,1.5);

\end{scope}

\end{tikzpicture}